\documentclass[journal]{IEEEtran}

\usepackage[noadjust]{cite}
\usepackage{subfigure}
\ifCLASSINFOpdf
   \usepackage[pdftex]{graphicx}
\else
   \usepackage[dvips]{graphicx}
\fi

\usepackage{amsmath}
\newtheorem{definition}{\indent Definition}
\newtheorem{example}{Example}
\newtheorem{proposition}{Proposition}

\usepackage{mathtools}

\usepackage[linesnumbered, ruled,vlined]{algorithm2e}
\SetKwRepeat{Do}{do}{while}
\usepackage{array}
\usepackage{dcolumn}
\usepackage[normalem]{ulem}
\useunder{\uline}{\ul}{}
\usepackage{multirow}
\usepackage{booktabs}

\usepackage{float}
\usepackage{changes}
\usepackage{url}
\usepackage{todonotes} % Load the todonotes package

\begin{document}

\title{Cascaded two-stage feature clustering and selection via separability and consistency in fuzzy decision systems}

\author{Yuepeng Chen,
        Weiping Ding,~\IEEEmembership{Senior Member,~IEEE,}
        Hengrong Ju,
        Jiashuang Huang,
        and Tao Yin% <-this % stops a space
\thanks{This paper is supported by the National Natural Science Foundation of China. (Corresponding author: Weiping Ding.)

Yuepeng Chen, Weiping Ding, Hengrong Ju, Jiashuang Huang, and Tao Yin are with the School of Information Science and Technology, Nantong University, Nantong 226019, China.

}% <-this % stops a space
% \thanks{Manuscript received April 19, 2005; revised August 26, 2015.}
}

% The paper headers
\markboth{IEEE TRANSACTIONS ON FUZZY SYSTEMS}%
{Shell \MakeLowercase{\textit{et al.}}: Bare Demo of IEEEtran.cls for IEEE Journals}

\maketitle

\begin{abstract}
Feature selection is a vital technique in machine learning, as it can reduce computational complexity, improve model performance, and mitigate the risk of overfitting. However, the increasing complexity and dimensionality of datasets pose significant challenges in the selection of features. 
Focusing on these challenges, this paper proposes a cascaded two-stage feature clustering and selection algorithm for fuzzy decision systems. In the first stage, we reduce the search space by clustering relevant features and addressing inter-feature redundancy. In the second stage, a clustering-based sequentially forward selection method that explores
the global and local structure of data is presented. We propose a novel metric for assessing the significance of features, which considers both global separability and local consistency. Global separability measures the degree of intra-class cohesion and inter-class separation based on fuzzy membership, providing a comprehensive understanding of data separability. Meanwhile, local consistency leverages the fuzzy neighborhood rough set model to capture uncertainty and fuzziness in the data. The effectiveness of our proposed algorithm is evaluated through experiments conducted on 18 public datasets and a real-world schizophrenia dataset. The experiment results demonstrate our algorithm's superiority over benchmarking algorithms in both classification accuracy and the number of selected features. 
\end{abstract}


\IEEEpeerreviewmaketitle

\section{Introduction}

\IEEEPARstart{W}{ith} the advent of the digital era, there has been an unprecedented surge in data from various sources such as sensors, social media, financial systems, and healthcare resources. However, traditional methods struggle to handle big data due to its high dimensionality, noise, and redundant information, significantly impacting the accuracy and efficiency of both data analysis and decision-making processes. 
To address these challenges, feature selection emerges as a crucial technique across pattern recognition, machine learning, and data mining. It involves selecting the most relevant features from the original set to prevent overfitting, enhance interpretability, and optimize learning task performance \cite{Ding2020multiple, ApplicationsLiu, ApplicationsZhao, ApplicationsYuan}.

In real-world scenarios, datasets often exhibit fuzziness and uncertainty due to high dimensions and substantial noise. While rough set theory \cite{I_Pawlak1982rough} as a valuable mathematical tool for feature selection has proven effective in handling uncertain information in classification problems, it faces limitations with continuous data. Numerous generalized models have emerged to address this issue, with fuzzy rough sets \cite{I_Dubois_fuzzy} playing a significant role in overcoming the limitations. 

Dubois and Prade \cite{I_Dubois_fuzzy} integrated the theories of rough sets and fuzzy sets, defining the fuzzy rough approximation operators and proposing the fuzzy rough set model, effectively addressing uncertainty and fuzziness in data analysis. 
Scholars have proposed various extended models based on the fuzzy rough sets.
Dai et al. \cite{I_FR_Dai_2018} introduced the concept of reduced maximal discernibility pairs within the framework of the fuzzy rough set model. They subsequently developed algorithms for selecting reduced maximal discernibility pairs and weighted reduced maximal discernibility pairs.
Hu et al. \cite{I_FR_Hu_Kernelized} integrated kernel functions into fuzzy rough set models, introducing kernelized fuzzy rough sets to compute fuzzy memberships among samples for classification approximation and to assess attribute approximation quality.
Zhao et al. \cite{I_FR_zhao_fuzzy_2019} introduced a feature reduction approach employing fuzzy rough sets and presented a novel model capable of computing lower and upper approximations specifically for hierarchical class structures.
Zhang et al. \cite{I_FR_zhang_2020} proposed an incremental feature selection approach using information entropy based on the fuzzy rough set, devising an active approach by selecting representative instances and formulating an incremental mechanism.
Huang et al. \cite{I_FR_Huang_2023} proposed an incremental approach for hierarchical classification, encompassing a sibling strategy to minimize negative sample impact and incorporating incremental updates upon new sample arrival.
Wang et al. \cite{III_Wang2016} introduced the fuzzy neighborhood rough set model (FNRS), combining fuzzy rough set and neighborhood rough set theories. They established the dependency between fuzzy decisions and feature subsets, utilizing it to assess candidate feature importance, and subsequently developed a greedy forward selection algorithm.
SUN et al. \cite{Applications_Sun} applied fuzzy neighborhood rough sets in multi-label classification based on maximum relevance minimum redundancy (MRMR).
SUN et al. \cite{III_SunEntropy} combined fuzzy neighborhood rough set and multi-granulation concepts, utilizing fuzzy neighborhood pessimistic multi-granulation entropy as an evaluation criterion for feature importance.

These methods based on the fuzzy rough sets above aim to find an optimal feature subset while maintaining the dependency or the fuzzy positive region. In practice, evaluating the significance of each feature often involves employing a widely used greedy forward-searching algorithm. This algorithm iteratively constructs an optimal feature subset by adding features based on the significance. Nonetheless, as dataset dimensions expand, the execution time of this algorithm significantly increases, posing a challenge to its scalability.

To address the scalability challenges posed by the greedy forward-searching algorithm, alternative approaches such as feature selection based on clustering have been explored.
Clustering-based feature selection clusters similar features while maintaining dissimilarity between clusters, aiming to select representative features from each cluster. After selecting a feature during each iteration, the clustering-based method involves discarding other similar features within the cluster containing that particular feature. This method significantly reduces the overall complexity of the search.
Zhu et al. \cite{I_group_Zhu} introduced a sequential feature selection algorithm based on affinity propagation clustering, selecting features from each subcluster, and collecting all selected features together. 
Jensen et al. \cite{I_group_Jensen} introduced a novel approach within the framework of fuzzy-rough sets, using feature grouping to mitigate processing overhead in large datasets. Their method involves assessing the correlation coefficient between features, enabling the identification of cohesive groups. 
Chormunge and Jena \cite{I_group_Smita} proposed a new method that eliminated irrelevant features by the k-means clustering method and then selected non-redundant features by correlation measure from each cluster.  
Shang et al. \cite{I_group_Shang} introduced a feature selection technique involving redundancy-based feature grouping by hierarchical clustering, and selecting top-ranked features according to classification capacity.
Fannia et al. \cite{I_group_Fannia} introduced an unsupervised feature selection algorithm, which integrates attribute clustering and rough set theory. Their approach adopted prototype clustering based on distance measurement to group similar attributes effectively.
Song et al. \cite{I_group_Song} introduced a feature deletion technique using a strategy guided by symmetric uncertainty correlation for feature clustering. They also enhanced the performance of the three phases by developing an improved PSO method.
These methods utilize feature similarities by leveraging clustering techniques to create feature groups, building upon assessing correlation coefficients among features.

Additionally, other clustering-based feature selection approaches investigate graph-based methods that exploit the relationships between features.
Song et al. \cite{I_group_SongQ} introduced a two-step feature selection algorithm utilizing graph-theoretic clustering to organize features into independent clusters, subsequently selecting the most relevant feature from each cluster. 
Zheng et al. \cite{I_group_Zheng} introduced a general framework based on graph theory and three-way mutual information for feature grouping. This framework operates by generating a minimum spanning tree based on the feature graph and clustering the features by iteratively removing edges from the resulting trees, from which representative features are subsequently selected.
Wan et al. \cite{GRMFS} introduced uncertainty measures within the fuzzy decision system to assess the interaction and redundancy between pairwise features within the graph structure. Feature subsets are obtained through the construction and decomposition of the minimum spanning tree based on the feature graph.
These methods utilize feature similarities by leveraging clustering techniques to create feature groups, building upon assessing correlation coefficients among features. As the dimensionality of data increases, the computational complexity of graph construction and clustering algorithms grows exponentially. Consequently,  the efficiency of these methods decreases when confronted with high-dimensional data.

This paper introduces a cascaded two-stage feature clustering and selection algorithm. The main contributions of this work are summarized as follows.

\begin{itemize}
\item
A cascaded two-stage feature clustering and selection framework is proposed to reduce the search space and address inter-feature redundancy by clustering the features into groups.
\end{itemize}

\begin{itemize}
\item
Based on the framework, a clustering-based sequentially forward selection algorithm is designed to select the features from feature groups, considering the interactions among unrelated features.
\end{itemize}

\begin{itemize}
\item
A fusion metric is proposed to evaluate the significance of features in fuzzy decision systems by assessing global separability and local consistency, where global separability evaluates intra-class cohesion and inter-class separation based on fuzzy membership, and local consistency captures uncertainty using the fuzzy neighborhood rough set model.
\end{itemize}

\begin{itemize}
\item
The paper performs experimental evaluations on public datasets and a real-world schizophrenia dataset. The results indicate that our approach performs better in terms of both the number of selected features and classification accuracy than six other feature selection techniques.
\end{itemize}

The structure of this article is as follows. Section \ref{I} provides the preliminaries, presenting the necessary background information. Section \ref{II} describes the two-stage feature clustering and selection algorithm. Section \ref{III} analyzes the experimental results of our proposed algorithm on real-world 18 datasets and the schizophrenia dataset. Section \ref{IV} presents the conclusions of this study.

\section{PRELIMINARIES} \label{I}
This section provides a brief overview of the fundamental concepts of the fuzzy C-Means (FCM) algorithm and fuzzy neighborhood rough set that are necessary to understand our work.

\subsection{Fuzzy C-Means algorithm}
The FCM algorithm \cite{FCM} is a clustering method that aims to partition a dataset into distinct groups according to the similarities between data points. It extends the classic k-means algorithm by incorporating fuzzy logic, allowing for a more flexible assignment of data points to clusters.

Let $X\mathrel{\mkern-4mu}=\mathrel{\mkern-4mu}\{x_1,x_2,\cdots,x_N\}$ be a dataset with $M$-dimensional samples.
The FCM algorithm operates by iteratively computing cluster centroids and a membership matrix to partition the data space effectively \cite{FHFS}. This process aims to minimize the following objective function
\begin{equation}
    J=\sum_{i=1}^{K} \sum_{k=1}^{N} u_{ik}^m \|x_k-v_i \| ^2
\end{equation}
where $K$ is the number of clusters, $u_{ik}$ denotes the membership degree of data $x_k$ to the $i$th cluster, $\|x_k-v_i \|$ is the Euclidean distance between $x_k$ and $v_i$ (the centroid of the $i$th cluster), and $m (m>1)$ is the weighting coefficient controlling the degree of fuzziness. The value of $m$ is conventionally set to 2.
Minimizing the objective function must satisfy a particular constraint \cite{FCM}
\begin{equation}
    \sum_{i=1}^{K}u_{ik}=1, 0 \le u_{ik} \le 1 \label{equation: objective function}
\end{equation}

Employing the Lagrange multiplier method, we can derive updated formulas for cluster centroids and a membership matrix. The algorithm can be summarized as follows. Initially, the cluster centroids $v = \{v_1,v_2,\cdots,v_c\}$ are initialized randomly. These centroids serve as the initial representatives for each cluster. Then, compute the degree of membership $u_{ik}$ for each data point $x_k$ to each cluster centroid $v_i$ using a fuzzy membership function
\begin{equation}
    u_{ik}=\frac{1}{\sum_{j=1}^{K} \left( \frac{\|x_k-v_i\|}{\|x_k-v_j\|} \right) ^{\frac{2}{m-1}}} \label{equation: membership}
\end{equation}

The objective function minimizes by allocating substantial membership values to input patterns closely located to their nearest cluster centers while assigning diminished membership values to those significantly distant from the cluster center.

Next, recalculate the cluster centroids based on the current memberships. The updated formula is defined as follows
\begin{equation}
    v_{i}=\frac{\sum_{k=1}^{N} u_{ik}^{m}x_{k}}{\sum_{k=1}^{N} u_{ik}^{m}} \label{equation: centroids}
\end{equation}

By iteratively updating memberships and centroids using equations (\ref{equation: membership}) and (\ref{equation: centroids}) to minimize the objective function, the FCM algorithm refines the cluster centers and memberships until convergence, yielding clusters that reflect the inherent structure within the dataset in a fuzzy manner. 
The iteration stops when $J^{(t)}-J^{(t-1)}< \epsilon$, where $J^{(t)}$ is the object function, $t$ is the number of iterations, and $\epsilon$ is a threshold given by the user.
At the end of the iterations, the final cluster centers $v$ represent the centroids of the clusters, and the membership matrix provides the degree of association of each data point to these clusters.

\subsection{Fuzzy neighborhood rough set}

Let $S\mathrel{\mkern-4mu}=\mathrel{\mkern-4mu} \langle U, A \cup D, f\rangle$ represent a fuzzy decision system, where $U = \left\{x_1,x_2,...,x_N\right\}$ denotes the universe or sample space, $A=\left\{a_1,a_2,...,a_M\right\}$ is the condition attributes or features used to characterize the samples, and $D$ represents the decision classes. Assuming the partitioning of samples into $c$ mutually exclusive decision classes by $D$, denoted as $U/D\mathrel{\mkern-7mu}=\mathrel{\mkern-7mu}\{ D_1, D_2,\cdots,D_c\}$, $f_a(x)$ represents the feature value of $x$ on condition attribute $a$. 
For any $x,y\in U$, $ B \subseteq A$, the feature subset $B$ can induce a fuzzy binary relation $R_B$ on $U$.
$R_B$ is the fuzzy similarity relation if it satisfies the reflexivity $R_B(x,x)=1$ and the symmetry $R_B(x,y)=R_B(y,x)$. 

In a fuzzy neighborhood decision system $S\mathrel{\mkern-4mu}=\mathrel{\mkern-4mu} \langle U, A \cup D, f, \lambda \rangle$, where $\lambda$ denotes the fuzzy neighborhood radius \cite{I_FR_wang_fitting}, the fuzzy neighborhood granule of any $x \in U$ based on $B$ is defined as
    
\begin{equation}
    \left[x\right]_B^\lambda(y)=\begin{cases}0, &R_B(x,y)<1-\lambda;\\
    R_B(x,y), &R_B(x,y)\geq 1-\lambda.
    \end{cases}
\end{equation}

In a fuzzy neighborhood decision system $S\mathrel{\mkern-4mu}=\mathrel{\mkern-4mu} \langle U, A \cup D, f, 
 \lambda \rangle$, $U/D\mathrel{\mkern-7mu}=\mathrel{\mkern-7mu}\{ D_1, D_2,\cdots,D_c\}$, the following is the  definition of the fuzzy decision of $x$ concerning $B$

\begin{equation}
    D_i(x)=\frac{\left|\left[x\right]_B^\lambda\cap D_i\right|}{\left|\left[x\right]_B^\lambda\right|}
\end{equation}
where $D_i$ is a fuzzy set and $D_i(x)$ indicates the membership degree of $x$ to $D_i$.

The fuzzy neighborhood lower and upper approximations of $D_i$ concerning $B$ \cite{I_FR_wang_fuzzy_2019} are indicated by

\begin{equation}
    \underline{R_B^\lambda}(D_i)=\{x \in D_i|[x]_B^\lambda(y)\subseteq D_i \}
\end{equation}

\begin{equation}
    \overline{R_B^\lambda}(D_i)=\{x \in D_i|[x]_B^\lambda(y)\cap D_i \neq \emptyset \}
\end{equation}

The positive region of $D$ concerning $B$ is expressed as follows
\begin{equation}
    POS_B^\lambda(D)=\bigcup_{i=1}^c\underline{R_B^\lambda} \left(D_i\right)
\end{equation}

The size of the positive region $POS_B^\lambda(D)$ serves as an indicative measure reflecting the classification ability inherent in the feature subset $B$.

In this study, we propose a fusion metric that integrates fuzzy membership and the fuzzy neighborhood rough set model to measure feature importance in fuzzy decision systems. The metric evaluates both global separability and local consistency: global separability assesses intra-class cohesion and inter-class separation using fuzzy membership, while local consistency captures uncertainty through the fuzzy neighborhood rough set model.

\section{The proposed method} \label{II}
\begin{figure*}[htbp]
    \centering
	\includegraphics[width=1\linewidth]{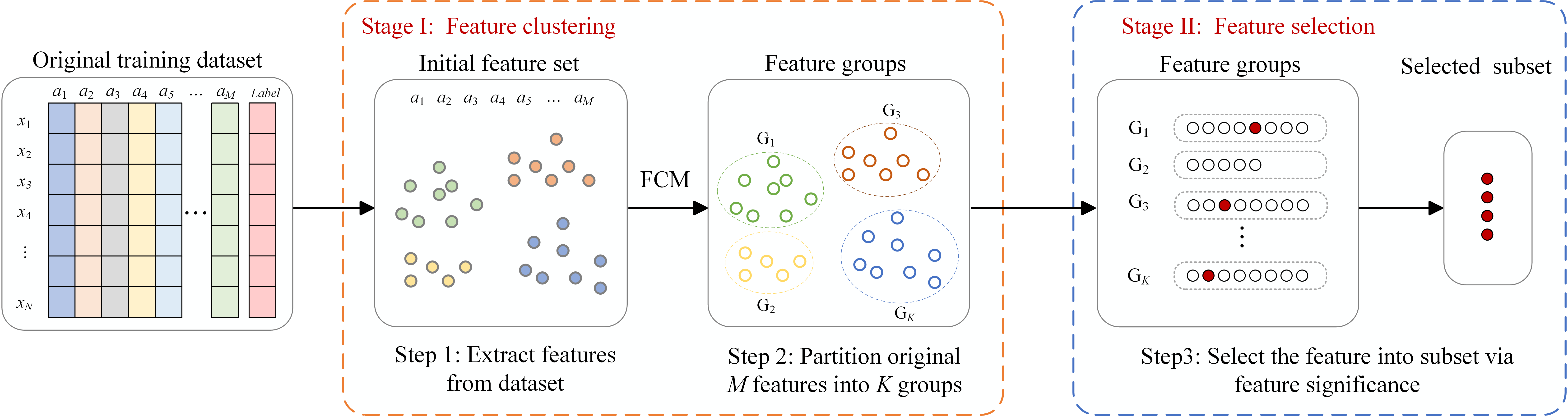}
	\caption{The framework of feature clustering and selection}
	\label{fig: framework}
\end{figure*}

This section presents a cascaded two-stage feature clustering and selection method, the framework of which is depicted in Figure \ref{fig: framework}.
The algorithm framework consists of two stages: feature clustering and feature selection. 
In the first stage, similarity-based clustering categorizes relevant features into multiple groups by FCM. Following this, the second stage employs a clustering-based sequentially forward selection algorithm, navigating through these feature clusters to select the subset of features. Based on the framework, we design a clustering-based sequentially forward selection algorithm to select the feature subset from feature groups.

\subsection{Feature clustering based on FCM}
Cluster analysis involves classifying data objects based on their similarities, to achieve low inter-cluster similarity and high intra-cluster similarity for effective clustering. By analyzing the correlation between features to conduct cluster analysis, redundant features can be grouped within the same cluster. In the clustering stage of our proposed method, features are regarded as data objects, and the Euclidean distance between feature elements is computed as a measure of similarity among features. Our approach primarily involves dividing the original features into distinct clusters, where the features within each cluster exhibit higher correlations, while the correlations between features from different clusters are comparatively lower.

\begin{definition}
Let $X\mathrel{\mkern-4mu}=\mathrel{\mkern-4mu}\{x_1,x_2,\cdots,x_N\}$ be a training set, where $x_n \in X$ is a sample vector comprising $M$ measurements from the feature set $A=\left\{a_1,a_2,\cdots,a_M\right\}$.
Feature clustering is the partitioning of a set $A$ into a collection $G =\{G_1,G_2,\cdots,G_K\}$ of mutually disjoint subsets of features, where $K$ is the number of feature clusters, with $A = G_1\cup G_2,\cdots \cup G_K$, and $G_1 \cap G_2 = \emptyset$.
\end{definition}

In the feature clustering stage, the FCM algorithm is employed. 
In the random initialization stage of FCM, there are multiple methods to choose initial cluster centers. If the initial cluster center selection is improper, it can lead to slow convergence or unbalanced cluster sizes. Therefore, we employ the KMeans++ method to initialize the clustering centers in FCM \cite{III_FCMIitial}.

\subsection{Feature selection based on global separability and local consistency}
\begin{figure*}[htbp]
    \centering
	\includegraphics[width=1\linewidth]{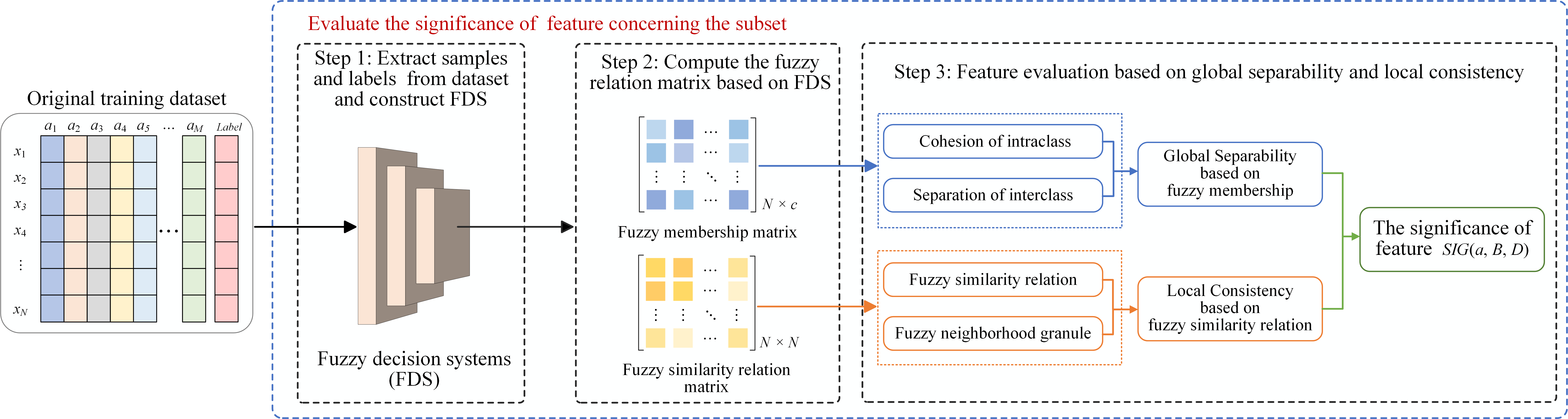}
	\caption{The framework for measuring the significance of features}
	\label{fig: framework2}
\end{figure*}

This section presents a fusion metric to measure the significance of features in fuzzy decision systems by fusing global separability and local consistency, illustrated in Figure \ref{fig: framework2}. 
The process involves several key steps. First, samples and labels are extracted from the dataset, and a fuzzy decision system is constructed. Subsequently, in the second step, the fuzzy membership matrix and fuzzy similarity matrix are computed based on the fuzzy decision system. In the third step, global separability assesses intra-class cohesion and inter-class separation by leveraging fuzzy membership, while local consistency captures uncertainty through the fuzzy neighborhood rough set model.

\subsubsection{Global separability based on fuzzy membership}
This section introduces two key measures, namely the degree of intra-class cohesion (DIC) and the degree of inter-class separation (DIS). These measures offer insights into the data structure, focusing on intra-class cohesion and inter-class separability, providing an intuitive understanding of the data structure.

The degree of intra-class cohesion (DIC) is proposed to evaluate the proximity among intra-class objects based on their fuzzy memberships within decision classes.
\begin{definition}
    Given a fuzzy decision system $S\mathrel{\mkern-4mu}=\mathrel{\mkern-4mu} \langle U, A \cup D, f \rangle$, where $U = \left\{x_1,x_2,...,x_N\right\}$, and $U/D\mathrel{\mkern-4mu}=\mathrel{\mkern-4mu}\{ D_1, D_2,\cdots,D_c\}$, $D_i(i=1,2,\cdots,c)$ is the $i$th decision class. The membership degree of sample $x_k$ concerning the decision class $D_i$ under feature subset $B(B \subseteq  A)$ is define as
    \begin{equation}
    u_B(x_k,D_i)=\frac{1}{\sum_{j=1}^{c} \left( \frac{\Delta_B(x_k,v_i)}{\Delta_B(x_k,v_j)} \right) ^{\frac{2}{m-1}}} \label{equation_u}
    \end{equation}
    
    \begin{equation}
    \Delta_B(x_k,D_i)=\sqrt[2]{\sum\limits_{a\in B}\left|f_a(x_k)-m_a(D_i)\right|^2}
    \label{equation_delta}
    \end{equation}
where $\Delta_B(x_k,D_i)$ denotes the Euclidean distance between object $x_k$ and the centroid $v_i$ of decision class $D_i$ under feature subset $B$, $f_a(x_k)$ is the feature value of $x_k$ on condition attribute $a$, and $m_a(D_i) = \frac{1}{|D_i|}{\sum_{x_k \in D_i} f_a(x_k) }$ denotes the mean value of the feature across samples belonging to the decision class $D_i$ under condition attribute $a$. $m$ is set to 2 to simplifies the calculation of $u_B(x_k,v_i)$ by reducing the square root and square operations.

The degree of intra-class cohesion (DIC) evaluates the compactness of intra-class objects by considering the memberships of samples to decision class, defined as
    \begin{equation}
        \mathrm{DIC}_B(D_i)=\frac{\sum_{x_k \in D_i} u_B(x_k,D_i)}{|D_i|}
    \end{equation}
\end{definition}
where $|D_i|$ is the cardinality of decision class $D_i$.
\begin{definition}
    Given a fuzzy decision system $S\mathrel{\mkern-4mu}=\mathrel{\mkern-4mu} \langle U, A \cup D, f \rangle$, $U/D\mathrel{\mkern-4mu}=\mathrel{\mkern-4mu}\{D_1, D_2,\cdots,D_c\}$. For feature subset $B(B \subseteq A)$, the degree of intra-class cohesion (DIC) for the fuzzy decision system concerning $B$ is defined as follows
    \begin{equation}
        \mathrm{DIC}_B(D) = \frac{1}{c}\sum_{i=1}^{c} \mathrm{DIC}_B(D_i)
    \end{equation}
\end{definition}

The DIC represents the collective membership degree of samples within the feature subset concerning their corresponding decision classes. As shown in (\ref{equation_u}), it clarifies a negative correlation between membership and distance, wherein a larger DIC indicates a closer proximity among intra-class objects. Therefore, a higher DIC within the feature subset $B$ signifies increased intra-class compactness among the samples.

\begin{proposition}
    In a fuzzy decision system $S$, $D_i \subseteq U/D$, $B \subseteq A$, then $0 \le \mathrm{DIC}_B(D) \le 1$.
\end{proposition}

The measure of separability within a fuzzy system relies not only on the cohesion of intra-class objects but also on the dispersion of inter-class objects. To address the comprehensive evaluation, we propose a validity index DIS to measure inter-class separation by utilizing the distance measure between fuzzy sets. To calculate the distance between fuzzy sets, we employ the similarity measure proposed by Lee et al. \cite{III_SimilarityMB}.

\begin{definition}
Let $S\mathrel{\mkern-4mu}=\mathrel{\mkern-4mu} \langle U, A \cup D, f \rangle$ be a fuzzy decision system, $U/D\mathrel{\mkern-4mu}=\mathrel{\mkern-4mu}\{D_1, D_2,\cdots,D_c\}$, and $B \subseteq A$. The similarity of two decision classes $D_i$ and $D_j$ at $x_k$ under $B$ is defined as
\begin{equation}
    S_B(D_i,D_j:x_k) = \mathrm{min}(u_B(x_k,D_i), u_B(x_k,D_j))
\end{equation}
\end{definition}

The similarity between two decision classes is defined as a weighted summation of $S_B(D_i,D_j:x_k) $ for all samples in $U$. The similarity between two decision classes $D_i$ and $D_j$ under $B$ is defined as
\begin{equation}
    S_B(D_i,D_j) = \frac{c}{N}\sum_{k=1}^{N} w(x_k)S_B(D_i,D_j:x_k) 
    \label{eq: S_B}
\end{equation}

where $w(x_k)$ represents the weight of overlapping data points, depending on the uncertainty of overlapping data points between decision classes, defined as
\begin{equation}
    w(x_k)= \frac{H(x_k)}{\mathrm{max}_{1\le k \le N}{H(x_k)}}
    \label{eq: weight}
\end{equation}
where $H(x_k)= -\sum_{i=1}^{c} u_B(x_k,D_i) \log{u_B(x_k,D_i)}$ is the entropy of sample $x_k$.

The degree of inter-class separation (DIS) is the average similarity value among pairs of clusters, defined as
\begin{equation}
    \mathrm{DIS}_B(D) = 1-\frac{2}{c(c-1)} \sum_{i \neq j}S_B(D_i,D_j)
    \label{eq: DIS}
\end{equation}

\begin{proposition}
    In a fuzzy decision system $S$, $D_i \subseteq U/D$, $B \subseteq A$, then $0 \le \mathrm{DIS}_B(D) \le 1$.
\end{proposition}

% \begin{proof}
% $u_B(x_k,D_i)$ is the fuzzy membership of the sample $x_k$ belonging to the decision class $D_i$, we have $0 \le u_B(x_k,D_i) \le 1$. Then, we can obtain $0 \le \mathrm{min}(u_B(x_k,D_i),u_B(x_k,D_j)) \le \frac{1}{c}$ and $0<w(x_k) \le 1$. Thus, we can get $0 \le S_B(D_i,D_j) \le 1$. We can get the mean of $S_B(D_i,D_j)$ between each pair of decision classes from 0 to 1. 
% The degree of inter-class separation under $B$ satisfies $0 \le \mathrm{DIS}_B(D) \le 1$.
% \end{proof}

\begin{definition}
Let $\mathrm{DIC}_B(D)$ and $\mathrm{DIS}_B(D)$ represent the degrees of cohesion and separation concerning $B$. The global separability of the fuzzy decision system concerning $B$ is defined as
    \begin{equation}
        \mathrm{GS}_B(D) = \mathrm{DIC}_B(D) \cdot \mathrm{DIS}_B(D)
        \label{eq: gs}
    \end{equation}
\end{definition}

\begin{proposition}
    Assuming $S$ a fuzzy decision system, $B \subseteq A$, $\mathrm{GS}_B(D)$ is the global separability of the fuzzy decision system concerning $B$, then $0 \le \mathrm{GS}_B(D) \le 1$.
\end{proposition}
    
% \begin{proof}
%     It follows directly from propositions 1 and 2.
% \end{proof}

\subsubsection{Local consistency based on fuzzy similarity relation }
The labeling consistency between adjacent samples within the feature subset is presumed to correlate with their similarity in labels, thereby indicating the identification capability of the feature subset. Wang et al. \cite{III_Wang2016} proposed the FNRS-based feature selection algorithm, yielding significant classification outcomes. Within the model, $\lambda$ serves as a parameter governing the fuzzy neighborhood radius's scale, while $\alpha$ regulates the degree of inclusion, mitigating the influence of noise in the data. The two parameters should be selected for the suitable value. Inspired by the concept in \cite{III_SunIF}, a fusion of expert knowledge regarding features with empirical experience is employed to dynamically adjust the knowledge threshold, thereby establishing the fuzzy neighborhood similarity relation as follows.

\begin{definition}
    Let $S\mathrel{\mkern-4mu}=\mathrel{\mkern-4mu} \langle U, A \cup D, f \rangle$ be a fuzzy decision system, where $U\mathrel{\mkern-6mu} =\mathrel{\mkern-6mu} \left\{x_1,x_2,...,x_N\right\}$. $\forall a \in A$ and $x,y \in U$, the fuzzy similarity relation between $x$ and $y$ on feature $a$ is defined as
    \begin{equation}
        R_a(x,y)=\begin{cases}0, &\mathrm{if} |f_a(x)-f_a(y)|>\delta_a\\
        1-|f_a(x)-f_a(y)|, &\mathrm{if} |f_a(x)-f_a(y)|\leq \delta_a
    \end{cases}
    \end{equation} 
where $\delta_{a}$ is the adaptive fuzzy neighborhood radius, can be calculated as $\delta_{a} = \frac{\sigma(a)}{\pi}$. $\sigma(a)$ represents the standard deviation under the feature $a$ and $\pi$ controls the size of the fuzzy neighborhood of samples. 
\end{definition}

The adaptive neighborhood fuzzy granule $\left[x\right]_a \left(y\right)$ is induced by the adaptive fuzzy relation $R_a$.
Adaptive neighborhoods allow for the generation of fuzzy granules of varied sizes, integrating both the data's distribution information $\sigma(a)$ and the constraint parameter $\pi$. The neighborhood adaptive fuzzy granule of $x$, induced by the feature subset $B$, is determined as $\left[x\right]_B \left(y\right)=\mathrm{min}_{a\in B} \left\{ \left[x\right]_a\left(y\right) \right\}$.

\begin{definition}
Assuming $S\mathrel{\mkern-4mu}=\mathrel{\mkern-4mu} \langle U, A \cup D, f \rangle$, $U/D = \{D_1,\cdots,D_c\}$. $\forall x_i \in U$, $L(x_i)$ represents the set of samples with the same labels as $x_i$. The local consistency for fuzzy neighbor samples on the feature subset $B$ is defined as
\begin{equation}
    \mathrm{LC}_B(D) = \frac{1}{|U|} \sum_{x_i \in U}\frac{|\left[x_i\right]_B \cap L(x_i)|}{|\left[x_i\right]_B|}
\end{equation}
where $|\left[x_i\right]_B| = \sum_{j=1}^N R_B(x_i,x_j)$ represents the cardinality of the fuzzy granule induced by feature subset $B$.
\end{definition}

\begin{example}
   Given a fuzzy decision system $S\mathrel{\mkern-4mu}=\mathrel{\mkern-4mu} \langle U, A \cup D, V, f \rangle$, where $U = \left\{x_1,x_2,...,x_6\right\}$, $A=\left\{a_1,a_2,\cdots,a_M\right\}$. Assume that $U/D=\{D_1,D_2\}$, such that $D_1=\{x_1,x_2,x_3\}$ and $D_2=\{x_4,x_5,x_6\}$. $R_B$ is the fuzzy similarity relation induced by $B(B\subseteq A)$, as depicted below
   
\begin{equation*}
R_B=
\begin{bmatrix}
1 & 0.6 & 0.8 & 0 & 0 & 0 \\
0.6 & 1 & 0.9 & 0.5 & 0 & 0 \\
0.8 & 0.9 & 1 & 0.7 & 0.6 & 0 \\
0 & 0.5 & 0.7 & 1 & 0 & 0.8 \\
0 & 0 & 0.6 & 0 & 1 & 0.7 \\
0 & 0 & 0 & 0.8 & 0.7 & 1 \\
\end{bmatrix}
\end{equation*}

Then we can get $[x_1] = \frac{1}{x_1} + \frac{0.6}{x_2} + \frac{0.8}{x_3} + \frac{0}{x_4} + \frac{0}{x_5} + \frac{0}{x_6}$ and $L(x_1)= D_1 = \{x_1,x_2,x_3\}$. Thus $|\left[x_1\right]_B|=|\left[x_i\right]_B \cap L(x_i)|=1+0.6+0.8=2.4$. And we can get $\mathrm{LC}_B(D) =\frac{1}{6} \times (\frac{2.4}{2.4} + \frac{2.5}{3} + \frac{2.7}{4} + \frac{1.8}{3} + \frac{1.7}{2.3} + \frac{2.5}{2.5}) = 0.81$.
\end{example}

\subsubsection{Feature selection via global separability and local consistency}
In this section, the fusion of global separability and local consistency forms a feature evaluation function, which quantifies the discriminatory capacity of features. The function is defined as follows:

\begin{definition}
    In a $S\mathrel{\mkern-4mu}=\mathrel{\mkern-4mu} \langle U, A \cup D, f \rangle$, $B \subseteq A$, the discrimination ability degree of $D$ regarding $B$ is defined as
    \begin{equation}
        \mathlarger{\gamma}_B(D) = \beta \cdot \mathrm{GS}_B(D) + (1-\beta) \cdot \mathrm{LC}_B(D)
        \label{eq: lc}
    \end{equation}
\end{definition}
where $\beta$ $(0 \leq \beta \leq 1)$ is a tuned parameter to balance the significance of global separability and local consistency.

\begin{proposition}
    Assuming $\mathrm{GS}_B(D)$ and $\mathrm{LC}_B(D)$ are the global separability and local consistency on the feature subset $B$, respectively. When $0 \leq \mathrm{GS}_B(D) \leq 1$ and $0 \leq \mathrm{LC}_B(D) \leq 1$, and $0 \leq \beta \leq 1$, then $0 \leq \mathlarger{\gamma}_B(D)\leq 1$ and a larger $\mathlarger{\gamma}_B(D)$ represents the stronger discriminative ability of $B$.
\end{proposition}

Then, we establish the feature significance concerning a feature subset, a crucial step in our feature selection method.

\begin{definition}
    Given a $S\mathrel{\mkern-4mu}=\mathrel{\mkern-4mu} \langle U, A \cup D, f \rangle$, $B \subseteq A$, $a \in A-B$ the significance of the feature $a$ concerning the feature subset $B$ is defined as
    \begin{equation}
        SIG(a,B,D) = \mathlarger{\gamma}_{B\cup \{a\}}(D)-\mathlarger{\gamma}_{B}(D)
        \label{eq: sig}
    \end{equation}
\end{definition}
    
$SIG(a, B, D)$ represents a metric quantifying the significance of feature $a$ concerning $B$ within decision $D$.

\begin{algorithm}[htbp]
\SetCommentSty{normalfont}
\small
%\normalsize
\SetAlgoLined %显示end
\LinesNumbered%显示行号
\caption{Feature clustering and selection via global separability and local consistency (FCSSC) algorithm} \label{algorithm 1}
\KwIn{A fuzzy decision system $S\mathrel{\mkern-4mu}=\mathrel{\mkern-4mu} \langle U, A \cup D, V, f \rangle$,  number of clusters $K$, threshold $\delta$.}
\KwOut{Feature subset $reduct$.}
Initialize $reduct \Leftarrow \emptyset$ \;
Divide features into $K$ groups $G = \left\{G_1,G_2,\cdots,G_K\right\}$ by FCM\;
\While{$G \neq \emptyset$}{
    \If{$|reduct| \geq \delta$}{break;}
    $sig \Leftarrow -1$\;
    $index \Leftarrow None$\;
    \For{\rm{each} $a_i\in G$}{
    $tmp = a_i \cup reduct$\;
    Compute the global separability of $tmp$ by (\ref{eq: gs})\;
    Compute the local consistency of $tmp$ by (\ref{eq: lc})\;
    Compute the significance of feature $a_i$ with respect to $reduct$ $SIG(a_i, reduct, D)$ by (\ref{eq: sig})\;
    \If{$SIG(a_i, reduct, D) > max_{sig}$}{
    $sig \Leftarrow SIG(a_i, reduct, D)$\;
    $index \Leftarrow i$\;
    }
    }
    $reduct \Leftarrow reduct\cup a_{index}$\;
    $G \Leftarrow G-G_j$ \;  \tcp{Delete the feature set $G_j$ containing feature $a_{index}$.}
}
\textbf{return} $reduct$\;
\end{algorithm}
Based on the previous definition, we design a feature clustering and selection via global separability and local consistency (FCSSC) algorithm. The feature clustering and selection process is summarized in Algorithm \ref{algorithm 1}. 
Firstly, we divide the conditional attributes into $K$ groups, denoted as $G = \left\{G_1,G_2,\cdots,G_K\right\}$, by FCM.
Then, the algorithm begins by initializing an empty subset $reduct$ to store the selected features. 
Each feature $a_i$ within the set $G$ is assessed for global separability, local consistency, and the significance of feature $a_i$ concerning $reduct$. Subsequently, the feature $a_i$ that produces the highest value of $SIG(a_i,B,D)$ is then added into the feature subset $reduct$. Simultaneously, the feature set $G_j$ containing the selected feature $a_{index}$ is removed from the feature group $G$. This iterative process continues, updating the feature groups $G$ and the $reduct$, until either the feature group $G$ becomes empty or the cardinality of $reduct$ exceeds the termination threshold $\delta$.

In our FCSSC algorithm, we set the number of clusters as $K= \lceil \sqrt{M} \log{M} \rceil$, where $M$ is the count of condition attributes. Assuming the number of features in each group is $l_1,l_2,\cdots,l_K$, $l_1 \leq l_2 \leq \cdots \leq l_K$, and $K \geq \delta$. In the worst scenario, the total number of searches for features in our approach is reduced compared to conventional heuristic searches. Precisely, the count of searches within our method can be computed as $M + (M-l_1) + (M-l_1-l_2) + \cdots + 1$. By structuring clusters based on the correlation of the features, our algorithm efficiently reduces the search space, resulting in a considerably reduced number of required searches compared to traditional heuristic approaches.

\section{Experimental Analysis} \label{III}
To validate the effectiveness of our proposed method, we conduct experiments comparing it with six other feature selection algorithms: CMIM \cite{CMIM}, mRMR \cite{mRMR}, ReliefF \cite{ReliefF}, FNRS \cite{III_Wang2016}, GRMFS \cite{GRMFS}, and FHFS \cite{FHFS}.

Assessment of the feature selection results involves three established classifiers: $K$-Nearest Neighborhood (KNN), support vector machine (SVM), and classification and regression tree (CART), representing diverse supervised classification algorithms. Classification accuracy for the selected feature subsets is determined using default parameter settings for these classifiers. Employing a ten-fold cross-validation method, our experiments entail the random partitioning of the original dataset into ten subsets, nine for training and one for testing, repeated ten times to compute average and standard deviation values, constituting the final results. Our method is evaluated and compared using the same training and testing datasets as other feature selection algorithms. 

All feature selection algorithms are implemented in Python, and executed on hardware equipped with Intel Core i7-10870H CPU \makeatletter@\makeatother 2.20 GHz and 16 GB RAM. 

\subsection{Experiment on the UCI and the Kent Ridge Biomedical Datasets}
For this study on feature selection, 18 real-world datasets from the UCI repository of machine learning databases \cite{UCIRepository} and the Kent Ridge biomedical data set \cite{Repository}, encompass diverse fields such as medical diagnosis, image segmentation, and texture classification, shown in Table \ref{tab: Data description}. Missing values in datasets are imputed utilizing the maximum probability value approach. To mitigate variability resulting from differences in magnitudes, the maximum–minimum method normalizes the numerical data range to $[0, 1]$.

\begin{table}[htbp]
\setlength{\abovecaptionskip}{0cm} % 调整间距
\setlength{\belowcaptionskip}{-0.2cm}
\centering
\caption{Data set description.}
\label{tab: Data description}  
\begin{tabular}{ccccc}
    \toprule
    No. & Datasets   & Samples & Features & Classes  \\
    \midrule
    1   &Abalone    & 4177    & 8          & 3        \\
    2   &Climate    & 540     & 18         & 2        \\
    3   &Credit     & 690     & 15         & 2        \\
    4   &Diabetes   & 768     & 8          & 2        \\
    5   &Ionosphere & 351     & 33         & 2        \\
    6   &Pima       & 768     & 8          & 2        \\
    7   &Seeds      & 210     & 7          & 3        \\
    8   &Segment    & 2310    & 18         & 7        \\
    9   &Sonar      & 208     & 60         & 2        \\
    10   &Speaker    & 329     & 12         & 6        \\
    11   &Wdbc       & 569     & 31         & 2        \\
    12   &Wine       & 178     & 14         & 3        \\
    13   &Wpbc       & 198     & 34         & 2        \\
    14   &Zas        & 303     & 21         & 2        \\
    15   &Tumors     & 327     & 12558      & 7        \\
    16   &DLBCL      & 77      & 7130       & 2        \\
    17   &MLL        & 72      & 12582      & 3        \\
    18   &Prostate   & 136     & 12600      & 2        \\
    \bottomrule
\end{tabular}
\end{table}

In our FCSSC algorithm, there are two parameters $\beta$ and $\delta$. The parameter $\beta$, balancing the significance of global separability and local consistency, is set from 0 to 1 in increments of 0.1. Moreover,  we do not employ the first stage of clustering in low-dimensional datasets. Regarding the termination parameter $\delta$, for the initial fourteen low-dimensional datasets, $\delta$ aligns with the count of conditional attributes. Conversely, for the remaining four high-dimensional datasets, a uniform setting of $\delta = 50$ is employed. Similar termination parameter settings are applied to CMIM, mRMR, ReliefF, and FHFS, aligning with the strategy used in FCSSC. For FNRS, this algorithm employs distinct parameters: $\lambda$, influencing the size of the fuzzy neighborhood, which spans from 0.1 to 0.5 in increments of 0.05. Additionally, the parameter $\alpha$, determining the inclusion degree, ranges from 0.5 to 1 in steps of 0.05. GRMFS introduces two parameters: $\zeta$, managing the fuzzy granular size, which ranges from 0.2 to 2 in increments of 0.2. Furthermore, the parameter $\beta$ in GRMFS, affecting the exclusion of samples in different classes, varies from 0 to 1 in steps of 0.2.

\begin{table*}[htbp]
\centering
\setlength{\abovecaptionskip}{0cm} % 调整间距
\setlength{\belowcaptionskip}{-0.2cm}
\caption{Comparison of average classification accuracy under KNN (mean ± std).}
\label{TABLE: KNN}
\resizebox{\linewidth}{!}{
\begin{tabular}{cllllllll}
\toprule
Datasets   & RAW             & CMIM               & mRMR                    & ReliefF          & FNRS            & GRMFS                   & FHFS                    & FCSSC                       \\
\midrule
Abalone    & 0.6260 ± 0.0170 & 0.6157 ± 0.0114    & 0.5436 ± 0.0203         & 0.6068 ± 0.0231 & 0.5638 ± 0.0232 & 0.5805 ± 0.0193         & 0.5446 ± 0.0234         & \textbf{0.6256 ± 0.0188}  \\
Climate    & 0.9314 ± 0.0351 & 0.9203 ± 0.0234    & 0.9203 ± 0.0234         & 0.9480 ± 0.0182 & 0.9093 ± 0.0130 & \textbf{0.9480 ± 0.0200} & 0.9444 ± 0.0203         & 0.9426 ± 0.0193          \\
Credit     & 0.8753 ± 0.0324 & 0.6995 ± 0.0467    & 0.8390 ± 0.0515         & 0.7344 ± 0.0149 & 0.4957 ± 0.0600 & 0.8521 ± 0.0532         & 0.8551 ± 0.0686         & \textbf{0.8783 ± 0.0385}  \\
Diabetes   & 0.7339 ± 0.0575 & 0.7300 ± 0.0557    & 0.6362 ± 0.0391         & 0.7118 ± 0.0382 & 0.6796 ± 0.0390 & 0.7209 ± 0.0348         & 0.7329 ± 0.0578         & \textbf{0.7551 ± 0.0028}  \\
Ionosphere & 0.8514 ± 0.0636 & 0.8543 ± 0.0432    & 0.8657 ± 0.0496         & 0.8800 ± 0.0524 & 0.9201 ± 0.0380 & 0.8771 ± 0.0572         & 0.8830 ± 0.0352         & \textbf{0.9288 ± 0.0013}  \\
Pima       & 0.7419 ± 0.0554 & 0.7277 ± 0.0461    & 0.6285 ± 0.0537         & 0.7343 ± 0.0593 & 0.6731 ± 0.0381 & 0.7264 ± 0.0443         & 0.7329 ± 0.0578         & \textbf{0.7421 ± 0.0020}  \\
Seeds      & 0.9287 ± 0.0486 & 0.8469 ± 0.0731    & 0.8948 ± 0.0699         & 0.8755 ± 0.0682 & 0.7333 ± 0.0933 & 0.8474 ± 0.1079         & \textbf{0.9429 ± 0.0513} & \textbf{0.9429 ± 0.0415}  \\
Segment    & 0.9571 ± 0.0112 & 0.8480 ± 0.0246    & 0.9246 ± 0.0116         & 0.9034 ± 0.0224 & 0.9212 ± 0.0155 & 0.8809 ± 0.0204         & 0.9550 ± 0.0129         & \textbf{0.9641 ± 0.0124}  \\
Sonar      & 0.8074 ± 0.0776 & 0.7783 ± 0.0980    & 0.8067 ± 0.0678         & 0.8310 ± 0.0536 & 0.5088 ± 0.0918 & 0.8645 ± 0.0726         & 0.8552 ± 0.0949         & \textbf{0.8890 ± 0.0616}  \\
Speaker    & 0.7804 ± 0.0815 & 0.6157 ± 0.0324    & 0.6157 ± 0.0324         & 0.6069 ± 0.0519 & 0.5743 ± 0.0499 & 0.6343 ± 0.0459         & 0.6777 ± 0.0415         & \textbf{0.7993 ± 0.0370}  \\
Wdbc       & 0.9683 ± 0.0190 & 0.9295 ± 0.0253    & 0.9067 ± 0.0209         & 0.9313 ± 0.0244 & 0.7820 ± 0.0387 & 0.9260 ± 0.0334         & 0.9438 ± 0.0189         & \textbf{0.9772 ± 0.0228}  \\
Wine       & 0.9552 ± 0.0485 & 0.9268 ± 0.0439    & 0.8078 ± 0.0582         & 0.6663 ± 0.0477 & 0.8098 ± 0.0800 & 0.7163 ± 0.0716         & 0.9663 ± 0.0275         & \textbf{0.9778 ± 0.0276}  \\
Wpbc       & 0.7666 ± 0.1086 & 0.7513 ± 0.0597    & 0.7405 ± 0.0535         & 0.7361 ± 0.0931 & 0.7534 ± 0.0693 & 0.7311 ± 0.0908         & 0.7476 ± 0.0611         & \textbf{0.8092 ± 0.0573}  \\
Zas        & 0.7123 ± 0.0787 & 0.7053 ± 0.0575    & 0.7684 ± 0.0379         & 0.6848 ± 0.0585 & 0.6700 ± 0.0358 & 0.7426 ± 0.1005         & 0.7366 ± 0.0816         & \textbf{0.7955 ± 0.0568}  \\
Tumors     & 0.7524 ± 0.1360 & 0.7513 ± 0.0597    & \textbf{0.9500 ± 0.0764} & 0.9333 ± 0.0816 & 0.8714 ± 0.1238 & 0.9167 ± 0.0833         & 0.8690 ± 0.1243         & \textbf{0.9500 ± 0.0764}  \\
DLBCL      & 0.7850 ± 0.1704 & 0.7053 ± 0.0575    & \textbf{1.0000 ± 0}      & 0.9800 ± 0.0600 & 0.7690 ± 0.1243 & 0.9800 ± 0.0600         & \textbf{1.0000 ± 0}      & \textbf{1.0000 ± 0}       \\
MLL        & 0.8714 ± 0.1187 & \textbf{1.0000 ± 0} & \textbf{1.0000 ± 0}      & 0.9714 ± 0.0571 & 0.7881 ± 0.1323 & 0.9857 ± 0.0429         & 0.9607 ± 0.0821         & \textbf{1.0000 ± 0}       \\
Prostate   & 0.7852 ± 0.0699 & 0.9709 ± 0.0479    & 0.9709 ± 0.0479         & 0.9709 ± 0.0479 & 0.5676 ± 0.103  & 0.9407 ± 0.0556         & 0.9423 ± 0.0536         & \textbf{0.9852 ± 0.0297}  \\
Average    & 0.8239 ± 0.0683 & 0.7987 ± 0.0448    & 0.8233 ± 0.0397         & 0.817 ± 0.0485  & 0.7217 ± 0.0649 & 0.8262 ± 0.0563         & 0.8494 ± 0.0507         & \textbf{0.8872 ± 0.0272} \\
\bottomrule
\end{tabular}
}
\end{table*}

\begin{table*}[htbp]
\setlength{\abovecaptionskip}{0cm} % 调整间距
\setlength{\belowcaptionskip}{-0.2cm}
\centering
\caption{Comparison of average classification accuracy under SVM (mean ± std).}
\label{TABLE: SVM}
\resizebox{\linewidth}{!}{
\begin{tabular}{cllllllll}
\toprule
Datasets   & RAW             & CMIM               & mRMR                    & ReliefF          & FNRS            & GRMFS                   & FHFS                    & FCSSC                       \\
\midrule
Abalone    & 0.6466 ± 0.0208 & \textbf{0.6535 ± 0.0123} & 0.5579 ± 0.0223    & 0.6489 ± 0.0143         & 0.5858 ± 0.0269 & 0.6286 ± 0.0124    & 0.5700 ± 0.0183         & 0.6521 ± 0.0249          \\
Climate    & 0.9258 ± 0.0414 & 0.9240 ± 0.0209         & 0.9240 ± 0.0209    & 0.9517 ± 0.0265         & 0.9148 ± 0.0091 & 0.9526 ± 0.0190    & 0.9519 ± 0.0123         & \textbf{0.9537 ± 0.0148}  \\
Credit     & 0.8520 ± 0.0256 & 0.6560 ± 0.0444         & 0.8520 ± 0.0620    & 0.6560 ± 0.0444         & 0.5551 ± 0.0066 & 0.8550 ± 0.0653    & \textbf{0.8551 ± 0.0686} & \textbf{0.8551 ± 0.0547}  \\
Diabetes   & 0.7640 ± 0.0423 & 0.7510 ± 0.0385         & 0.6675 ± 0.0311    & 0.7432 ± 0.0436         & 0.7057 ± 0.0227 & 0.7575 ± 0.0419    & 0.7564 ± 0.0389         & \textbf{0.7708 ± 0.0261}  \\
Ionosphere & 0.9343 ± 0.0405 & 0.9371 ± 0.0379         & 0.9371 ± 0.0357    & \textbf{0.9514 ± 0.0405} & 0.9173 ± 0.0394 & 0.9314 ± 0.0366    & 0.9316 ± 0.0343         & 0.9458 ± 0.0287          \\
Pima       & 0.7653 ± 0.0552 & 0.7524 ± 0.0471         & 0.6716 ± 0.0468    & 0.7420 ± 0.0358         & 0.6967 ± 0.0287 & 0.7498 ± 0.0456    & 0.7564 ± 0.0389         & \textbf{0.7720 ± 0.0347}  \\
Seeds      & 0.9333 ± 0.0436 & 0.8374 ± 0.0882         & 0.8802 ± 0.0651    & 0.9043 ± 0.0738         & 0.7714 ± 0.0898 & 0.8617 ± 0.0861    & 0.9286 ± 0.0319         & \textbf{0.9524 ± 0.0014}  \\
Segment    & 0.9381 ± 0.0196 & 0.7549 ± 0.0307         & 0.7995 ± 0.0219    & 0.8224 ± 0.0207         & 0.9009 ± 0.0167 & 0.7402 ± 0.0260    & 0.9190 ± 0.0155         & \textbf{0.9338 ± 0.0082}  \\
Sonar      & 0.8405 ± 0.0651 & 0.7490 ± 0.0960         & 0.7340 ± 0.0979    & 0.8212 ± 0.0642         & 0.5767 ± 0.0829 & 0.8210 ± 0.0797    & 0.8269 ± 0.0863         & \textbf{0.8602 ± 0.0860}  \\
Speaker    & 0.7504 ± 0.0814 & 0.5762 ± 0.0434         & 0.5762 ± 0.0434    & 0.6401 ± 0.0634         & 0.5804 ± 0.0459 & 0.6616 ± 0.0674    & 0.6626 ± 0.0566         & \textbf{0.7507 ± 0.0651}  \\
Wdbc       & 0.9771 ± 0.0209 & 0.9172 ± 0.0262         & 0.9172 ± 0.0262    & 0.9190 ± 0.0239         & 0.8119 ± 0.0338 & 0.9154 ± 0.0308    & 0.9473 ± 0.0221         & \textbf{0.9754 ± 0.0210}  \\
Wine       & 0.9886 ± 0.0229 & 0.8922 ± 0.0654         & 0.8131 ± 0.0639    & 0.6781 ± 0.1148         & 0.8484 ± 0.0757 & 0.6892 ± 0.1158    & 0.9833 ± 0.0356         & \textbf{0.9889 ± 0.0222}  \\
Wpbc       & 0.7766 ± 0.1032 & 0.7618 ± 0.0231         & 0.7566 ± 0.0222    & 0.7566 ± 0.0222         & 0.7632 ± 0.0220 & 0.7566 ± 0.0222    & 0.7839 ± 0.0422         & \textbf{0.8145 ± 0.0614}  \\
Zas        & 0.7290 ± 0.0985 & 0.7119 ± 0.0145         & 0.7119 ± 0.0145    & 0.7119 ± 0.0145         & 0.7129 ± 0.0140 & 0.7217 ± 0.0234    & 0.7761 ± 0.0748         & \textbf{0.7858 ± 0.0174}  \\
Tumors     & 0.7833 ± 0.1833 & \textbf{0.9833 ± 0.0500} & 0.9667 ± 0.0667    & 0.9167 ± 0.0833         & 0.7631 ± 0.0576 & 0.9167 ± 0.0833    & 0.8690 ± 0.1243         & 0.9500 ± 0.0764          \\
DLBCL      & 0.8100 ± 0.1685 & \textbf{1.0000 ± 0}      & \textbf{1.0000 ± 0} & \textbf{1.0000 ± 0}      & 0.7525 ± 0.1124 & \textbf{1.0000 ± 0} & 0.9800 ± 0.0600         & \textbf{1.0000 ± 0}       \\
MLL        & 0.8589 ± 0.0905 & \textbf{1.0000 ± 0}      & 0.9857 ± 0.0429    & 0.9714 ± 0.0571         & 0.8393 ± 0.1412 & 0.9714 ± 0.0571    & 0.9607 ± 0.0821         & 0.9857 ± 0.0537          \\
Prostate   & 0.6742 ± 0.0753 & 0.9555 ± 0.0594         & 0.9632 ± 0.0488    & 0.9632 ± 0.0597         & 0.5440 ± 0.0265 & 0.9407 ± 0.0556    & 0.8176 ± 0.0721         & \textbf{0.9786 ± 0.0457}  \\
Average    & 0.8304 ± 0.0666 & 0.823 ± 0.0388          & 0.8175 ± 0.0407    & 0.8221 ± 0.0446         & 0.7356 ± 0.0473 & 0.8262 ± 0.0482    & 0.8487 ± 0.0508         & \textbf{0.8847 ± 0.0368}  \\
\bottomrule
\end{tabular}}
\end{table*}

Classification performance is typically used to evaluate the effectiveness of feature selection algorithms, with classification accuracy as the typical metric. To ensure robustness and reliability in our assessments, we mitigate the influence of computational randomness by averaging classification accuracy across multiple datasets. These aggregated results demonstrate the comparative effectiveness of each algorithm, shown in Tables \ref{TABLE: KNN}--\ref{TABLE: CART}. Conducting comprehensive comparisons involved ten rounds of calculations, leveraging average classification accuracy across three classifiers as benchmarks. The value with the highest classification performance is in bold, derived from mean and standard deviation values obtained from ten-fold cross-validation experiments, presented in the format of mean ± std. The final rows of the tables illustrate the average classification accuracy of these methods across all datasets, designated as `Average'.

By comparing Tables \ref{TABLE: KNN}--\ref{TABLE: CART}, we draw the following conclusions.
\begin{itemize}
\item
The proposed algorithm exhibited an average classification accuracy improvement of 6.33\%, 5.43\%, and 9.77\%, respectively, across the three classifiers compared to the original data. Moreover, FCSSC outperforms the other six algorithms in terms of both average classification accuracy and average standard deviation.
\end{itemize}
\begin{itemize}
\item
The proposed algorithm exhibits better performance than other feature selection algorithms on most data sets. With KNN as a classifier (Table \ref{TABLE: KNN}), our algorithm achieves the highest classification performance on 17 data sets. On the SVM classifier, CMIM, ReliefF, and FHFS achieve the highest classification accuracy on 4, 2, and 1 datasets, respectively.
The proposed method achieves the highest classification accuracy on 14 datasets with the SVM classifier. Similarly, FCSSC obtains the best results on 12 datasets with the CART classifier.
\end{itemize}
\begin{itemize}
\item
Across the results obtained from the three classifiers, it is evident that FCSSC stands out as highly effective and robust in feature selection for classification tasks.
\end{itemize}

\begin{figure*}[htbp]
\centering
\subfigure[Climate]{
\centering
\begin{minipage}[t]{0.3\textwidth}
\includegraphics[width=0.85\textwidth]{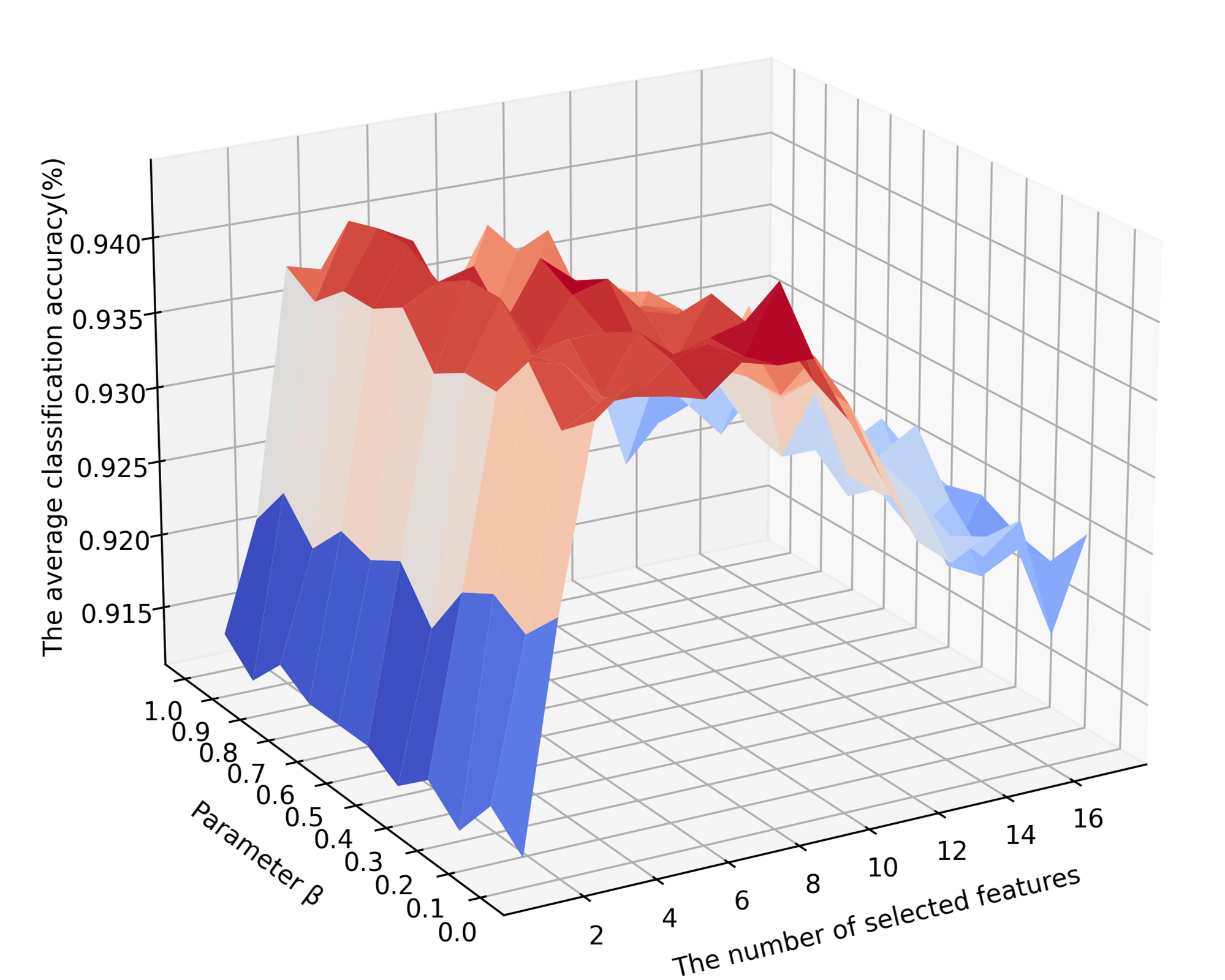}
\end{minipage}
}
\subfigure[Ionosphere]{
\centering
\begin{minipage}[t]{0.3\textwidth}
\includegraphics[width=0.85\textwidth]{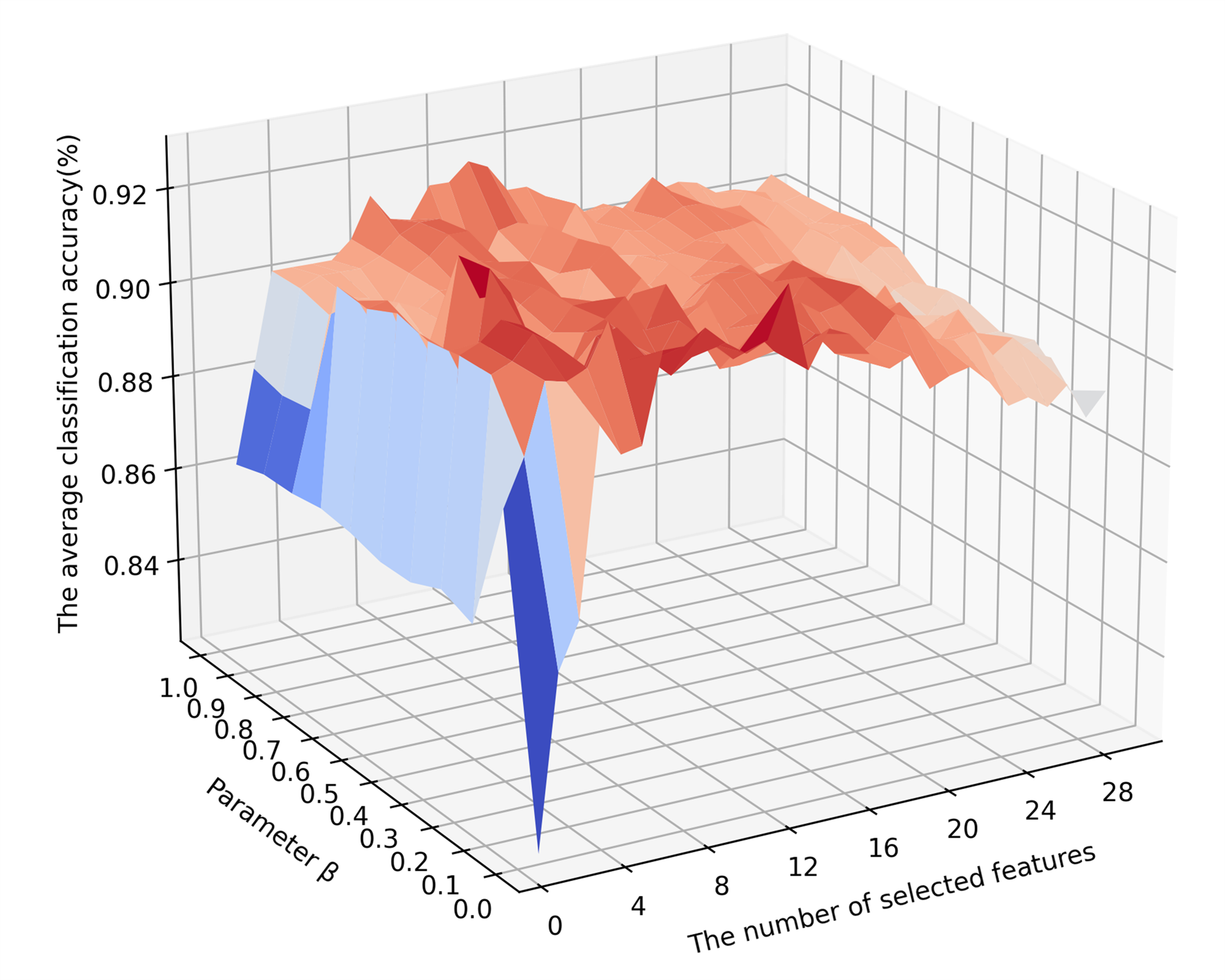}
\label{Segmentation}
\end{minipage}
}
\subfigure[Segment]{
\centering
\begin{minipage}[t]{0.3\textwidth}
\includegraphics[width=0.85\textwidth]{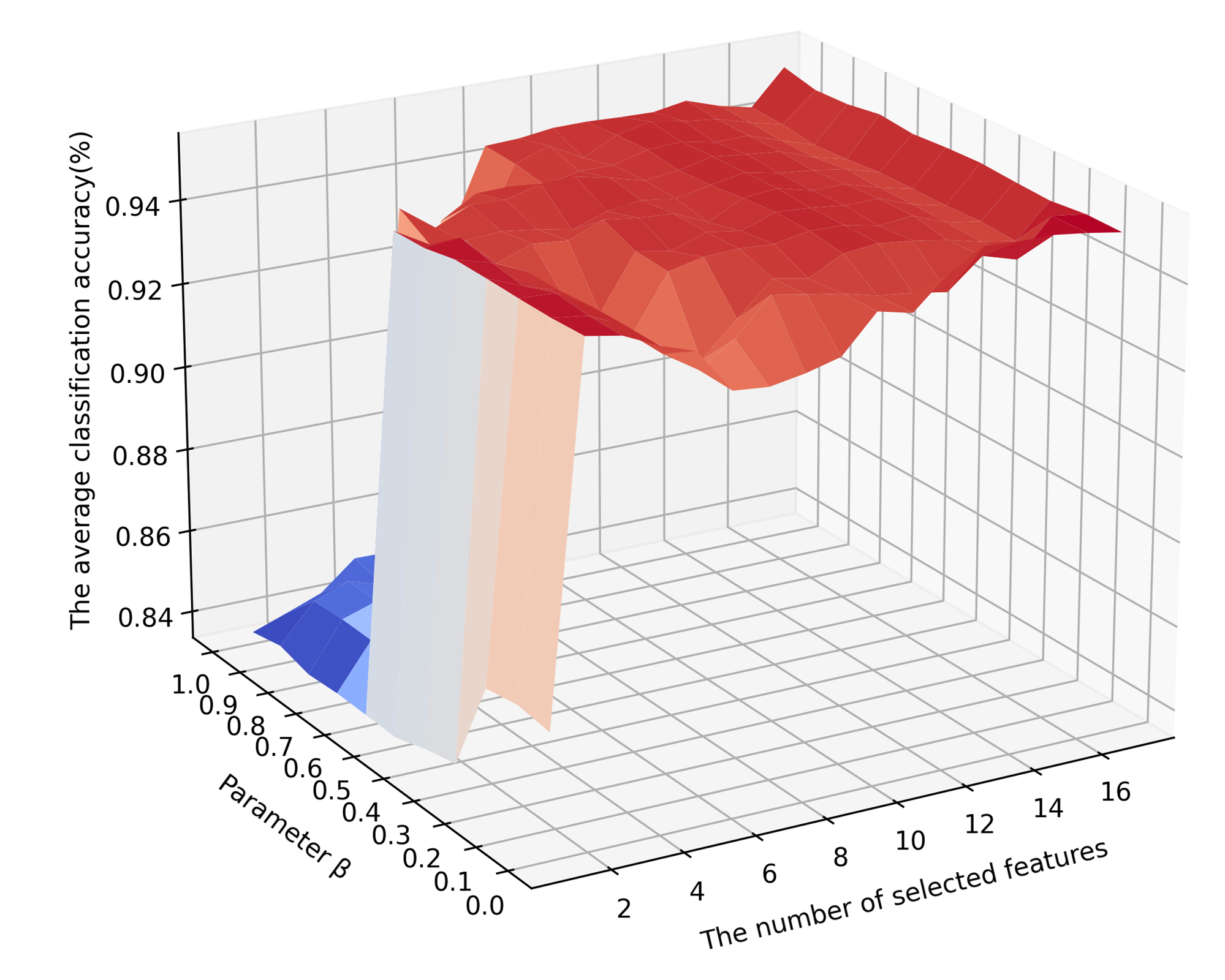}
\label{Spam}
\end{minipage}
}

\subfigure[Wine]{
\centering
\begin{minipage}[t]{0.3\textwidth}
\includegraphics[width=0.85\textwidth]{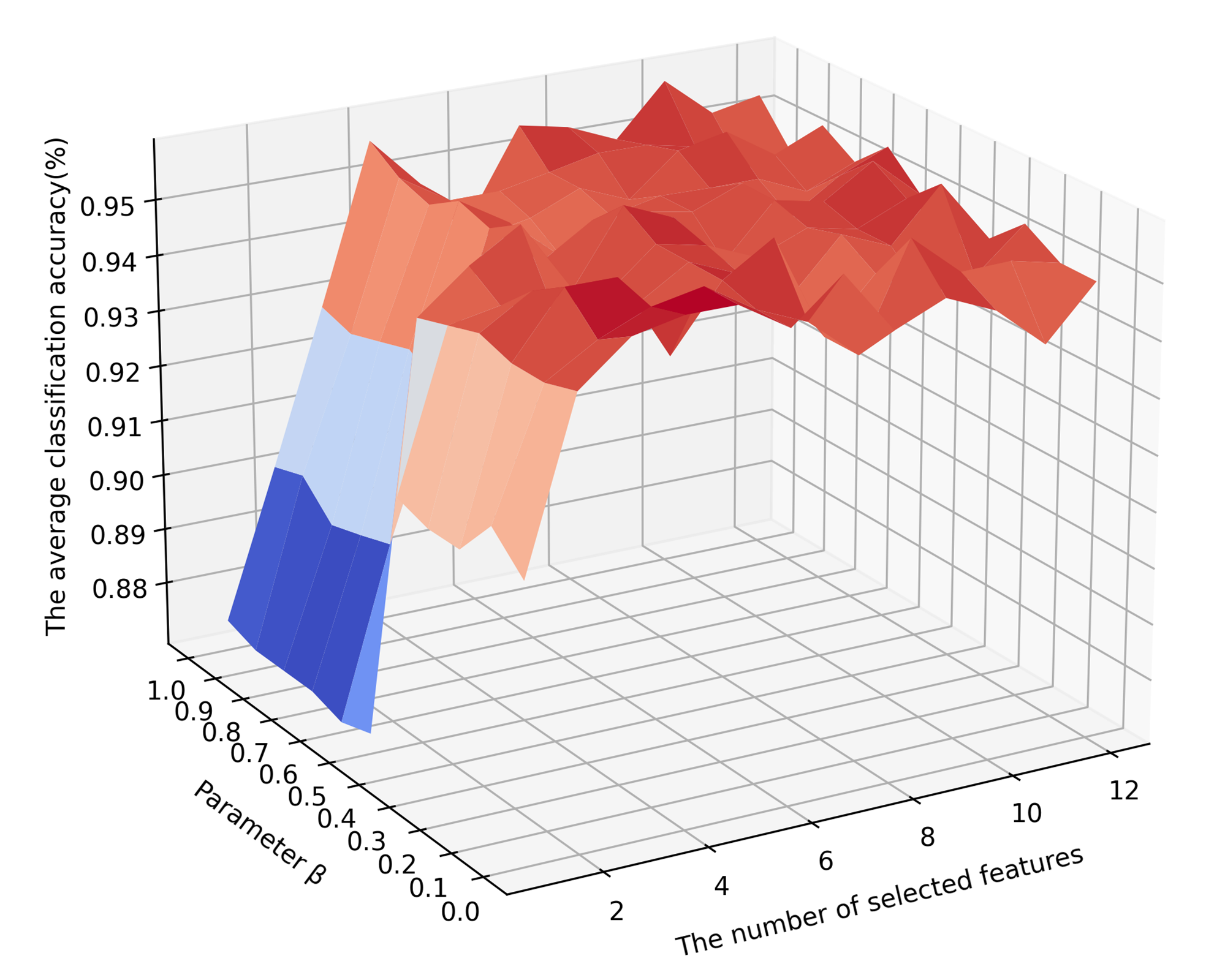}
\label{Texture}
\end{minipage}
}
\subfigure[DLBCL]{
\centering
\begin{minipage}[t]{0.3\textwidth}
\includegraphics[width=0.85\textwidth]{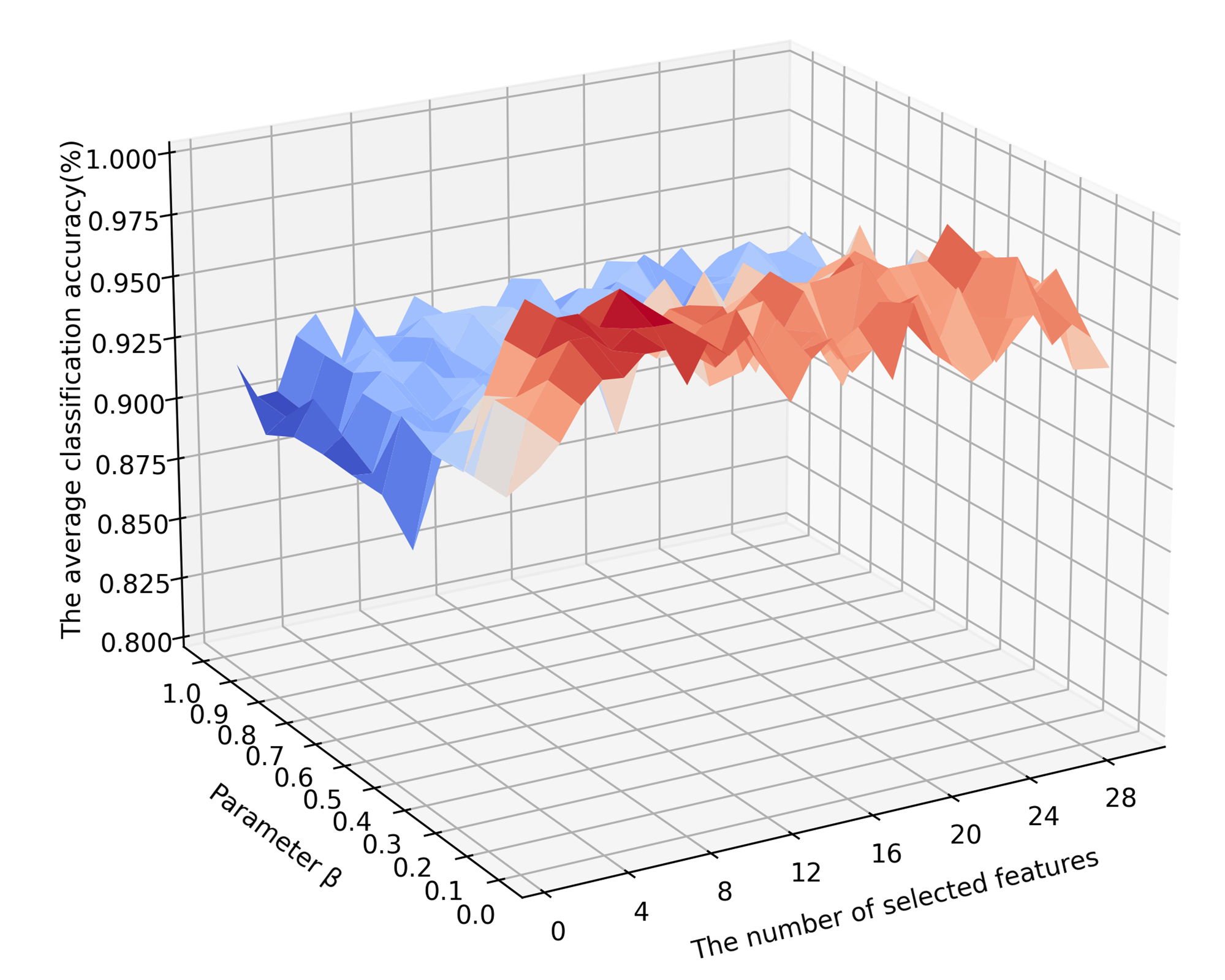}
\label{Waveform}
\end{minipage}
}
\subfigure[MLL]{
\centering
\begin{minipage}[t]{0.3\textwidth}
\includegraphics[width=0.85\textwidth]{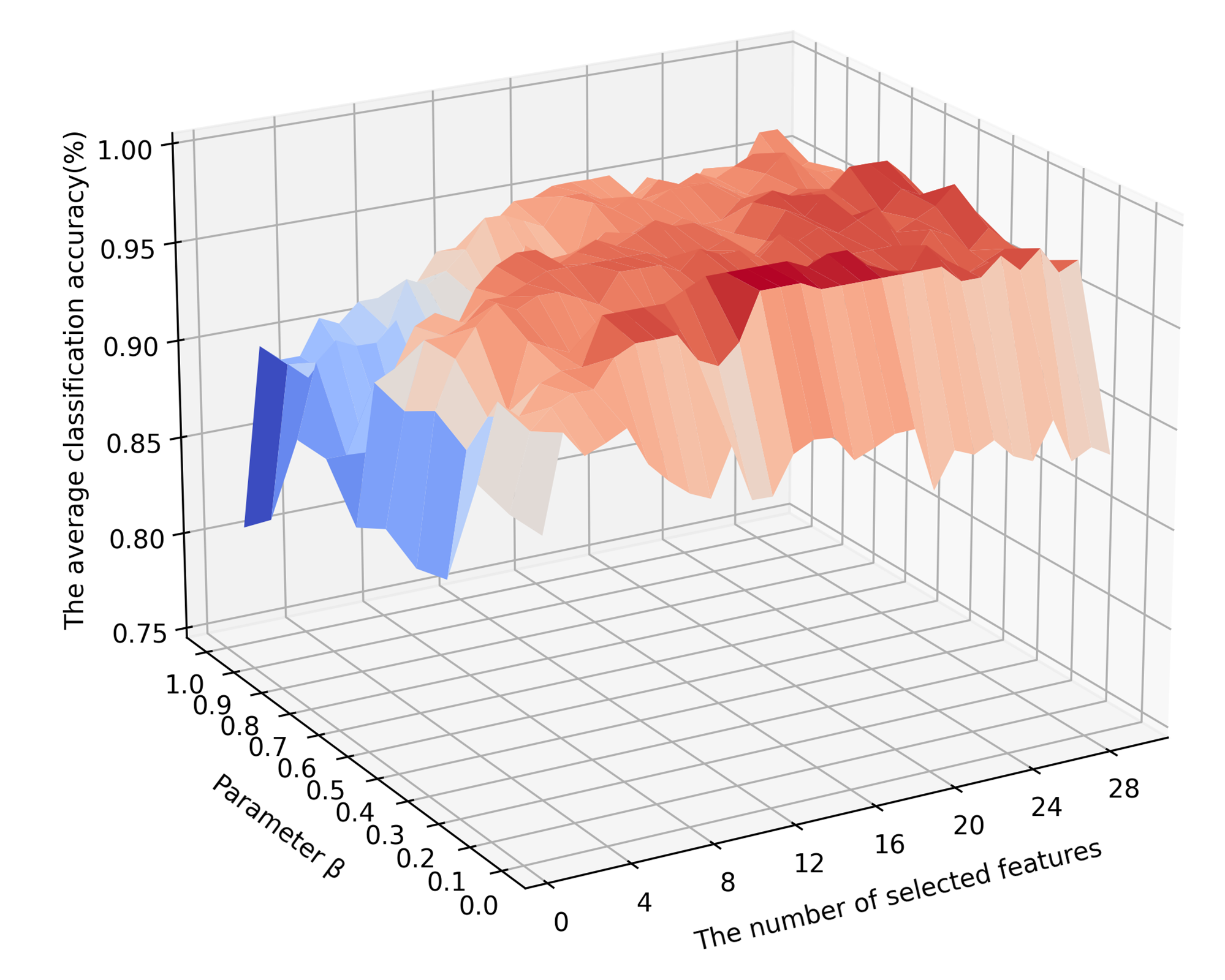}
\label{Wine}
\end{minipage}
}
\centering
\caption{The average classification accuracy variations with parameter $\beta$ and the number of features.}
\label{Fig: accuracy_para}
\end{figure*}

The algorithm's effectiveness undergoes further scrutiny by illustrating the average classification accuracy of three classifiers, clarifying its correlation with the parameter $\beta$ and the number of features (shown in Figure \ref{Fig: accuracy_para}). We varied the parameter $\beta$ from 0 to 1 with the step of 0.1, resulting in different numbers of selected features to achieve optimal classification performance. The x-axis, y-axis, and z-axis of the figure respectively represent the number of selected features, parameter $\beta$, and the average classification accuracy of the three classifiers. The analyses of Figure \ref{Fig: accuracy_para} illustrate the following conclusions.

Initially, the average classification accuracy in all data sets increases rapidly because the algorithm prioritizes highly differentiating features, specifically the first three. However, after this initial stage, the algorithm's performance stabilizes or experiences a slight decrease, as demonstrated by the trend changes on the data sets of Climate, Ionosphere, Segment, and Wine (Figures \ref{Fig: accuracy_para} (a), (b), (c), and (d)).
Furthermore, we note that the parameter $\beta$ affects the classification performance of some data sets, such as DLBCL and MLL (Figures \ref{Fig: accuracy_para} (e) and (f)). In contrast, for data sets such as Climate, Ionosphere, Segment, and Wine, the impact of parameter $\beta$ on the average classification performance of the three classifiers is relatively minor (Figures \ref{Fig: accuracy_para} (a), (b), (c), and (d)).
Notably, the highest average classification accuracy for the three classifiers varies with different values of the parameter $\beta$, leading to variations in the number of selected features for optimal classification accuracy. For example, the Segment dataset (Figure \ref{Fig: accuracy_para} (c)) achieves the highest average classification accuracy with three and six selected features when $\beta$ is set to 0 and 1, respectively.

\begin{table*}[htbp]
\setlength{\abovecaptionskip}{0cm} % 调整间距
\setlength{\belowcaptionskip}{-0.2cm}
\centering
\caption{Comparison of average classification accuracy under CART (mean ± std).}
\label{TABLE: CART}
\resizebox{\linewidth}{!}{
\begin{tabular}{cllllllll}
\toprule	
Datasets   & RAW             & CMIM               & mRMR                    & ReliefF          & FNRS            & GRMFS                   & FHFS                    & FCSSC                       \\
\midrule
Abalone    & 0.5788 ± 0.0221 & 0.5678 ± 0.0191         & 0.5187 ± 0.0300         & 0.5592 ± 0.0203         & 0.5274 ± 0.0223 & 0.5441 ± 0.0232         & \textbf{0.5719 ± 0.0264} & 0.5607 ± 0.0216          \\
Climate    & 0.8998 ± 0.0426 & 0.9055 ± 0.0449         & 0.9037 ± 0.0436         & 0.9222 ± 0.0318         & 0.8593 ± 0.0424 & 0.9258 ± 0.0233         & \textbf{0.9389 ± 0.0204} & \textbf{0.9389 ± 0.0275}  \\
Credit     & 0.8172 ± 0.0518 & 0.7968 ± 0.0447         & 0.8085 ± 0.0656         & 0.8143 ± 0.0693         & 0.5551 ± 0.0066 & 0.8550 ± 0.0653         & 0.8551 ± 0.0686         & \textbf{0.8609 ± 0.0519}  \\
Diabetes   & 0.7000 ± 0.0664 & 0.6687 ± 0.0430         & 0.5943 ± 0.0441         & 0.6635 ± 0.0507         & 0.6457 ± 0.0677 & 0.6920 ± 0.0824         & 0.7068 ± 0.0511         & \textbf{0.7147 ± 0.0616}  \\
Ionosphere & 0.9057 ± 0.0479 & 0.8943 ± 0.0339         & 0.8886 ± 0.0605         & 0.8943 ± 0.0572         & 0.8945 ± 0.0497 & 0.9171 ± 0.0632         & 0.9202 ± 0.0421         & \textbf{0.9230 ± 0.0480}  \\
Pima       & 0.6961 ± 0.0569 & 0.6807 ± 0.0431         & 0.5960 ± 0.0698         & 0.6599 ± 0.0493         & 0.6289 ± 0.0463 & 0.7133 ± 0.0462         & 0.7093 ± 0.0592         & \textbf{0.7238 ± 0.0448}  \\
Seeds      & 0.9233 ± 0.0439 & 0.8183 ± 0.0551         & 0.9186 ± 0.0567         & 0.8900 ± 0.0738         & 0.6905 ± 0.1112 & 0.8519 ± 0.0778         & 0.9286 ± 0.0488         & \textbf{0.9333 ± 0.0646}  \\
Segment    & 0.9636 ± 0.0093 & 0.9511 ± 0.0142         & 0.9407 ± 0.0085         & 0.9467 ± 0.0187         & 0.9260 ± 0.0198 & 0.9000 ± 0.0183         & 0.9641 ± 0.0097         & \textbf{0.9667 ± 0.0093}  \\
Sonar      & 0.7105 ± 0.0709 & 0.7286 ± 0.1013         & 0.7395 ± 0.0702         & \textbf{0.8405 ± 0.0577} & 0.5376 ± 0.1390 & 0.7969 ± 0.0975         & 0.7790 ± 0.0931         & 0.7595 ± 0.0864          \\
Speaker    & 0.6619 ± 0.0531 & 0.5304 ± 0.0534         & 0.5340 ± 0.0846         & 0.5337 ± 0.0611         & 0.4589 ± 0.0755 & 0.5735 ± 0.0537         & 0.5561 ± 0.0542         & \textbf{0.7325 ± 0.0604}  \\
Wdbc       & 0.9384 ± 0.0263 & \textbf{0.9436 ± 0.0313} & 0.9243 ± 0.0248         & 0.9278 ± 0.0228         & 0.7257 ± 0.0723 & 0.9260 ± 0.0270         & 0.9192 ± 0.0352         & 0.9402 ± 0.0141          \\
Wine       & 0.8987 ± 0.0595 & 0.8873 ± 0.0560         & 0.7850 ± 0.0663         & 0.9036 ± 0.0529         & 0.7144 ± 0.1050 & \textbf{0.9490 ± 0.0471} & 0.9275 ± 0.0557         & 0.9167 ± 0.0714          \\
Wpbc       & 0.6737 ± 0.0608 & 0.7008 ± 0.0766         & \textbf{0.7508 ± 0.0778} & 0.7139 ± 0.1021         & 0.6647 ± 0.0749 & 0.7092 ± 0.0859         & 0.6958 ± 0.0913         & 0.7124 ± 0.0851          \\
Zas        & 0.6227 ± 0.0586 & 0.6553 ± 0.0559         & 0.6491 ± 0.0738         & 0.7053 ± 0.0503         & 0.6933 ± 0.0333 & 0.7120 ± 0.0605         & 0.6965 ± 0.0455         & \textbf{0.7166 ± 0.0905}  \\
Tumors     & 0.7524 ± 0.1551 & \textbf{0.9667 ± 0.0667} & 0.9167 ± 0.0833         & 0.9000 ± 0.1106         & 0.7619 ± 0.2138 & 0.9000 ± 0.0816         & 0.9024 ± 0.0800         & 0.9024 ± 0.0800          \\
DLBCL      & 0.7300 ± 0.2182 & \textbf{1.0000 ± 0}      & \textbf{1.0000 ± 0}      & 0.9600 ± 0.1200         & 0.7405 ± 0.2631 & 0.9600 ± 0.1200         & 0.9400 ± 0.0917         & \textbf{1.0000 ± 0}       \\
MLL        & 0.8714 ± 0.1348 & 0.9286 ± 0.0958         & 0.9429 ± 0.0948         & 0.9429 ± 0.0948         & 0.8321 ± 0.0870 & 0.9429 ± 0.0700         & 0.9589 ± 0.0629         & \textbf{0.9714 ± 0.0571}  \\
Prostate   & 0.8132 ± 0.0961 & 0.9407 ± 0.0556         & 0.9110 ± 0.0642         & 0.9335 ± 0.0610         & 0.6022 ± 0.0893 & 0.9099 ± 0.0452         & 0.8192 ± 0.1164         & \textbf{0.9423 ± 0.0771}  \\
Average    & 0.7865 ± 0.0708 & 0.8092 ± 0.0495         & 0.7957 ± 0.0566         & 0.8173 ± 0.0614         & 0.6922 ± 0.0844 & 0.821 ± 0.0605          & 0.8216 ± 0.0585         & \textbf{0.8442 ± 0.0562} \\
\bottomrule
\end{tabular}}
\end{table*}

The size of the selected feature subset is a critical evaluation metric in feature reduction, reflecting the goal of achieving high classification accuracy with minimal features. Table \ref{Table: Comparison of numbers} illustrates the comparison of the number of selected features derived from the feature selection algorithms. Table \ref{Table: Comparison of numbers}, demonstrates that our algorithm consistently selects a significantly fewer average number of features compared to the raw dataset. In addition, our algorithm selects fewer features than those of the other methods.
This outcome strongly suggests our model's effectiveness in reducing feature dimensionality while preserving efficiency, outperforming other methods in feature selection.

\begin{table}[htbp]
\setlength{\abovecaptionskip}{0cm} % 调整间距
\setlength{\belowcaptionskip}{-0.2cm}
\centering
\caption{Comparison of numbers of selected features.}
\scriptsize
\label{Table: Comparison of numbers}
\resizebox{\linewidth}{!}{
\begin{tabular}{lcccccccc}
\toprule
Datasets   & RAW    & CMIM & mRMR & ReliefF & FNRS & GRMFS  & FHFS   & FCSSC  \\
\midrule
Abalone    & 8      & 6    & 3    & 3      & 6    & 2     & 2    & 4    \\
Climate    & 18     & 8    & 3    & 5      & 5    & 6     & 8    & 10   \\
Credit     & 15     & 2    & 5    & 4      & 1    & 2     & 2    & 7    \\
Diabetes   & 8      & 5    & 2    & 3      & 7    & 3     & 4    & 5    \\
Ionosphere & 33     & 19   & 11   & 9      & 6    & 10    & 13   & 10   \\
Pima       & 8      & 5    & 2    & 3      & 8    & 2     & 4    & 4    \\
Seeds      & 7      & 3    & 3    & 3      & 5    & 3     & 3    & 2    \\
Segment    & 18     & 10   & 7    & 7      & 10   & 7     & 4    & 3    \\
Sonar      & 60     & 14   & 24   & 9      & 12   & 17    & 14   & 14   \\
Speaker    & 12     & 8    & 5    & 3      & 11   & 5     & 5    & 10   \\
Wdbc       & 31     & 17   & 11   & 12     & 12   & 5     & 4    & 6    \\
Wine       & 14     & 10   & 2    & 5      & 6    & 5     & 5    & 6    \\
Wpbc       & 34     & 4    & 4    & 3      & 4    & 5     & 5    & 5    \\
Zas        & 21     & 3    & 7    & 7      & 5    & 3     & 3    & 9    \\
Tumors     & 12558  & 6    & 20   & 38     & 6    & 9     & 26   & 6    \\
DLBCL      & 7130   & 3    & 3    & 25     & 5    & 25    & 31   & 4    \\
MLL        & 12582  & 23   & 8    & 7      & 5    & 30    & 25   & 6    \\
Prostate   & 12600  & 30   & 40   & 49     & 6    & 29    & 3    & 6    \\
Average    & 2508.7 & 9.8  & 8.9  & 10.8   & 6.7  & 9.3   & 8.9  & 6.5  \\
\bottomrule
\end{tabular}}
\end{table}

Building on this evaluation, we conduct Friedman \cite{IV_Friedman} and Bonferroni-Dunn \cite{IV_post_Dunn} statistical tests to evaluate the classification performance of the seven feature selection algorithms. 
The Friedman test is a one-way repeated measures analysis of variance by ranks. Suppose that we have conducted $m$ approaches on $n$ data sets, and $r_i$ denotes the average rank of the $i$-th approach, then the Friedman statistic and its improved statistics are defined as follows

\begin{equation}
    \tau_{\chi^{2}}=\frac{12n}{m(m+1)}\left(\sum_{i=1}^{m}r_{i}^{2}-\frac{m(m+1)^{2}}{4}\right)
\end{equation}

\begin{equation}
\tau_F=\dfrac{(n-1)\tau_{\chi^2}}{n(m-1)-\tau_{\chi^2}}
\end{equation}

In the posthoc test \cite{IV_post_hoc}, the Friedman test's critical difference (CD) is computed as
$\mathrm{CD}_\alpha=q_\alpha\sqrt{\dfrac{m(m+1)}{6n}}$
where $q_\alpha$ is the critical value from the Studentized range distribution for a given significance level $\alpha$.

The Friedman test determines significant differences between approaches in different data sets. The critical difference, as a posthoc test, is employed to identify which approaches significantly differ from each other following a significant Friedman test. Upon computing the critical values for the Friedman statistics at significance levels of $\alpha=0.1$ and $\alpha=0.05$, which were $F(6,102) = 1.83$ and $F(6,102) = 2.19$, respectively, significant discrepancies are observed across the classifiers (KNN, SVM, CART), surpassing the established thresholds of $\alpha=0.1$ and $\alpha=0.05$ with values $\tau_{F}=10.15$, $10.48$, and $13.13$ for KNN, SVM, and CART classifiers, respectively.
Therefore, the null hypothesis, which assumes no difference between the approaches, is rejected. 
This rejection of the null hypothesis supports the existence of a notable difference in performance among the seven reduction algorithms under scrutiny. Subsequently, the Bonferroni-Dunn test revealed our algorithm's superior performance over all comparative methods, corroborating the statistical significance observed.

\subsection{Feature selection on the schizophrenia dataset}
After validating our algorithm's superiority through the evaluation of public datasets, our subsequent focus is to assess its real-world scalability, especially within the medical domain with a specific emphasis on the schizophrenia dataset.

The experiment is conducted utilizing the real fMRI dataset obtained from West China Hospital of Sichuan University \cite{V_Huang}. By meticulously processing this dataset and applying feature selection techniques, our objective is to improve the accuracy of schizophrenia prediction. This method facilitates a deeper exploration of the neurobiological basis of the disease and holds the promise of developing a more precise and tailored approach to diagnose and treat schizophrenia.
Utilizing resting-state functional magnetic resonance imaging (rs-fMRI) obtained via a 3-T General Electric MRI scanner, our preprocessing comprised multiple essential steps. These included initial data exclusion for stability, correction, spatial normalization, and filtering using SPM8 software and Data Processing Assistant \cite{V_Zhu}. Subsequently, we segmented the rs-fMRI images into 90 distinct brain regions utilizing the AAL template \cite{V_Functional}. Each region's time series data were derived from the mean value of all voxels within that specific area. This process enabled the representation of feature information for each brain region in the form of time series data. After data preprocessing, we obtained a dataset of 773 samples with 4005 connections between brain regions.

\begin{figure}[htbp]
\centering
\includegraphics[width=0.8\linewidth]{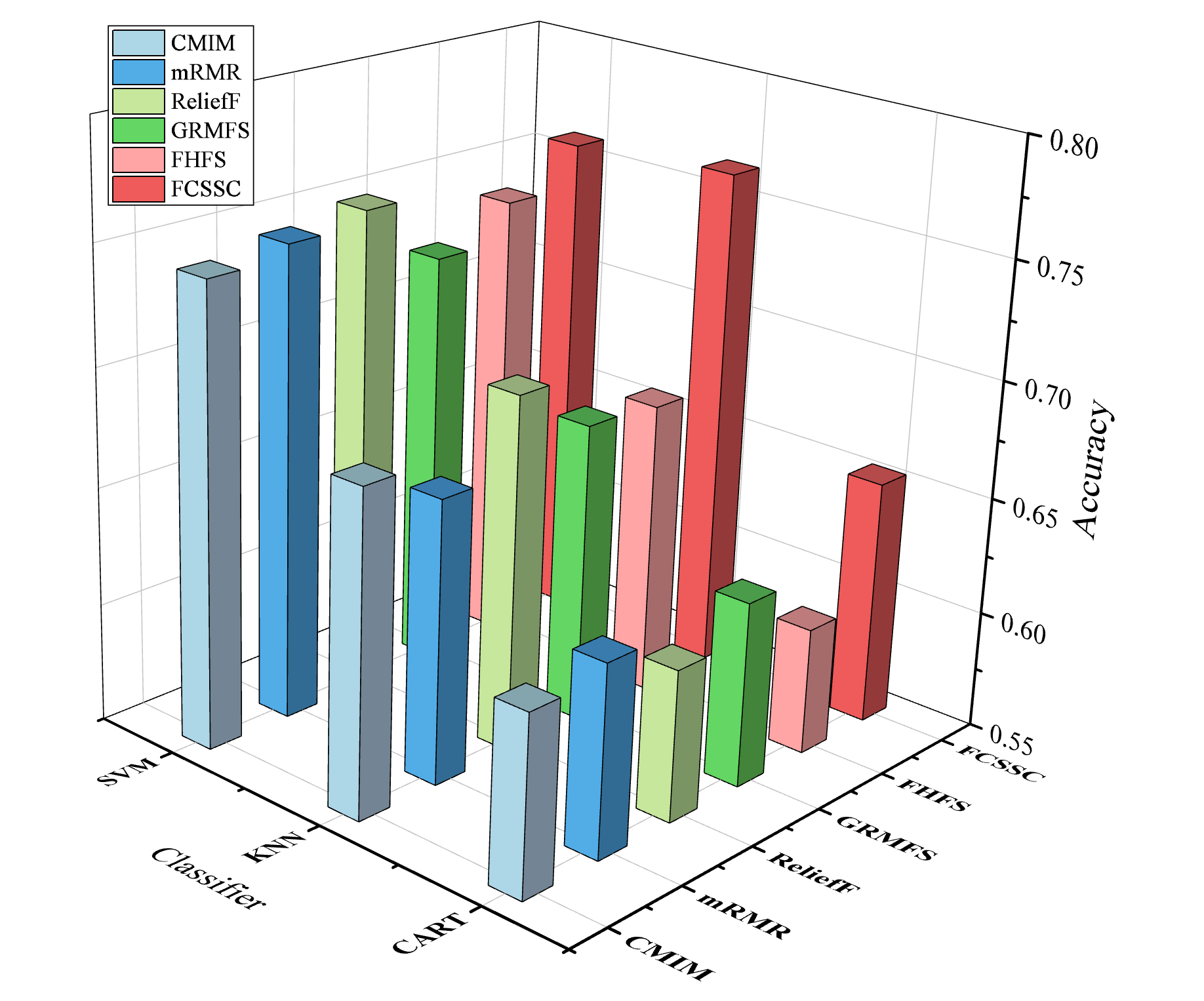}
\caption{The comparison of classification accuracy under three classifiers.}
\label{Fig: Brian_accuracy}
\end{figure}

The performance of our proposed approach is assessed on the schizophrenia dataset by evaluating its classification accuracy across various classifiers.
The KNN, SVM, and CART classifiers are employed for estimating the classification accuracy of these feature selection algorithms through 10-fold cross-validation. 
To visually depict the comparative performance, Figure \ref{Fig: Brian_accuracy} illustrates the classification accuracy achieved by the six feature selection algorithms under the three classifiers.
Our algorithm significantly improves accuracy from the original 0.5843 to 0.7683, outperforming the other comparative algorithms under the KNN classifier.
Under the SVM classifier, FCSSC achieved the highest classification accuracy of 0.7593, closely followed by ReliefF with 0.7552. Under the CART classifier, we observe that FCSSC attained the highest classification accuracy of 0.6541, outperforming the other five methods under the CART classifier.
These results collectively demonstrate the effectiveness and robustness of FCSSC for feature selection.

\begin{figure*}[htbp]
\centering
\subfigure[Accuracy]{
\centering
\begin{minipage}[t]{0.3\textwidth}
\includegraphics[width=\textwidth]{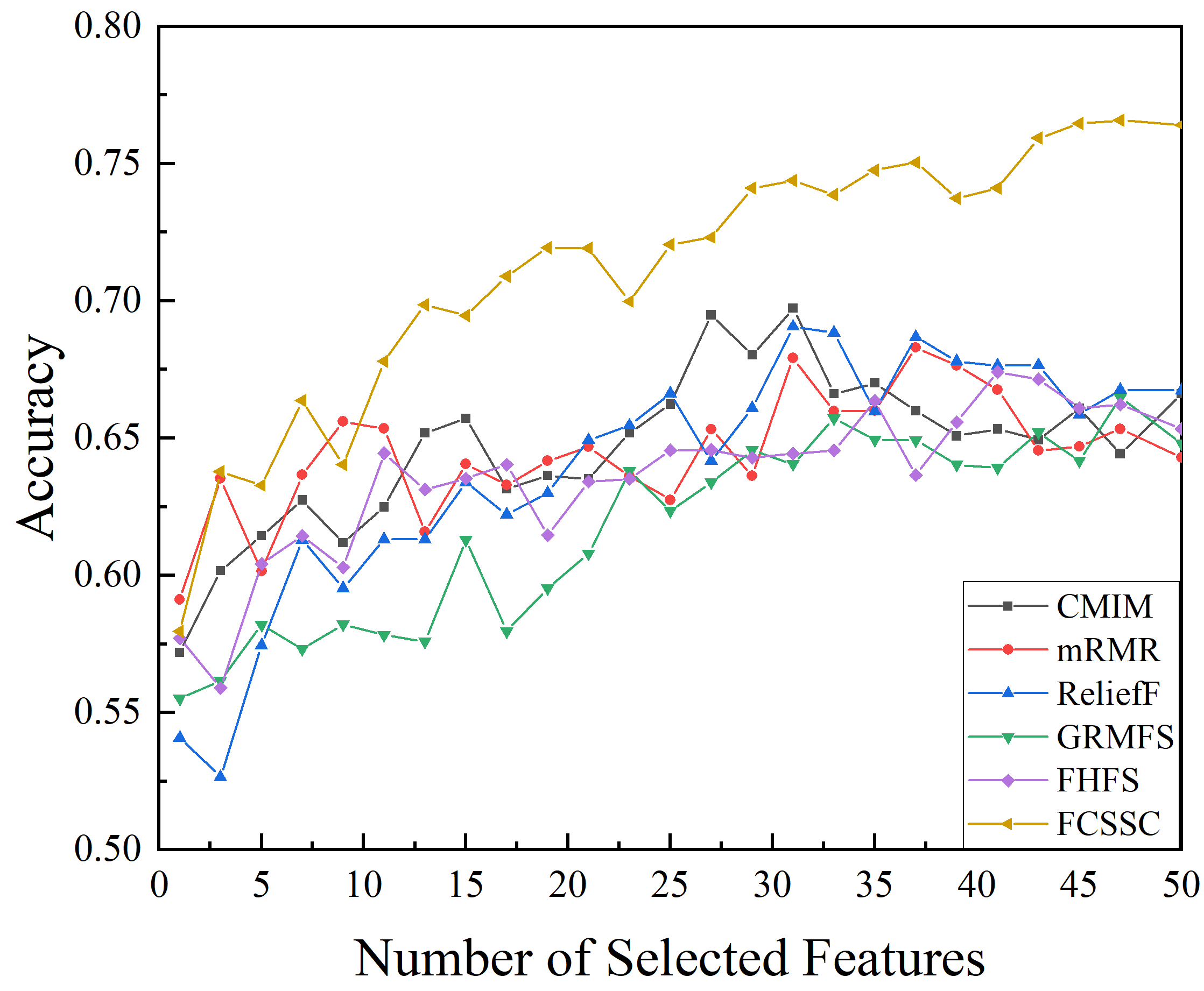}
\label{Fig: accuracy_features}
\end{minipage}
}
\subfigure[Precision]{
\centering
\begin{minipage}[t]{0.3\textwidth}
\includegraphics[width=\textwidth]{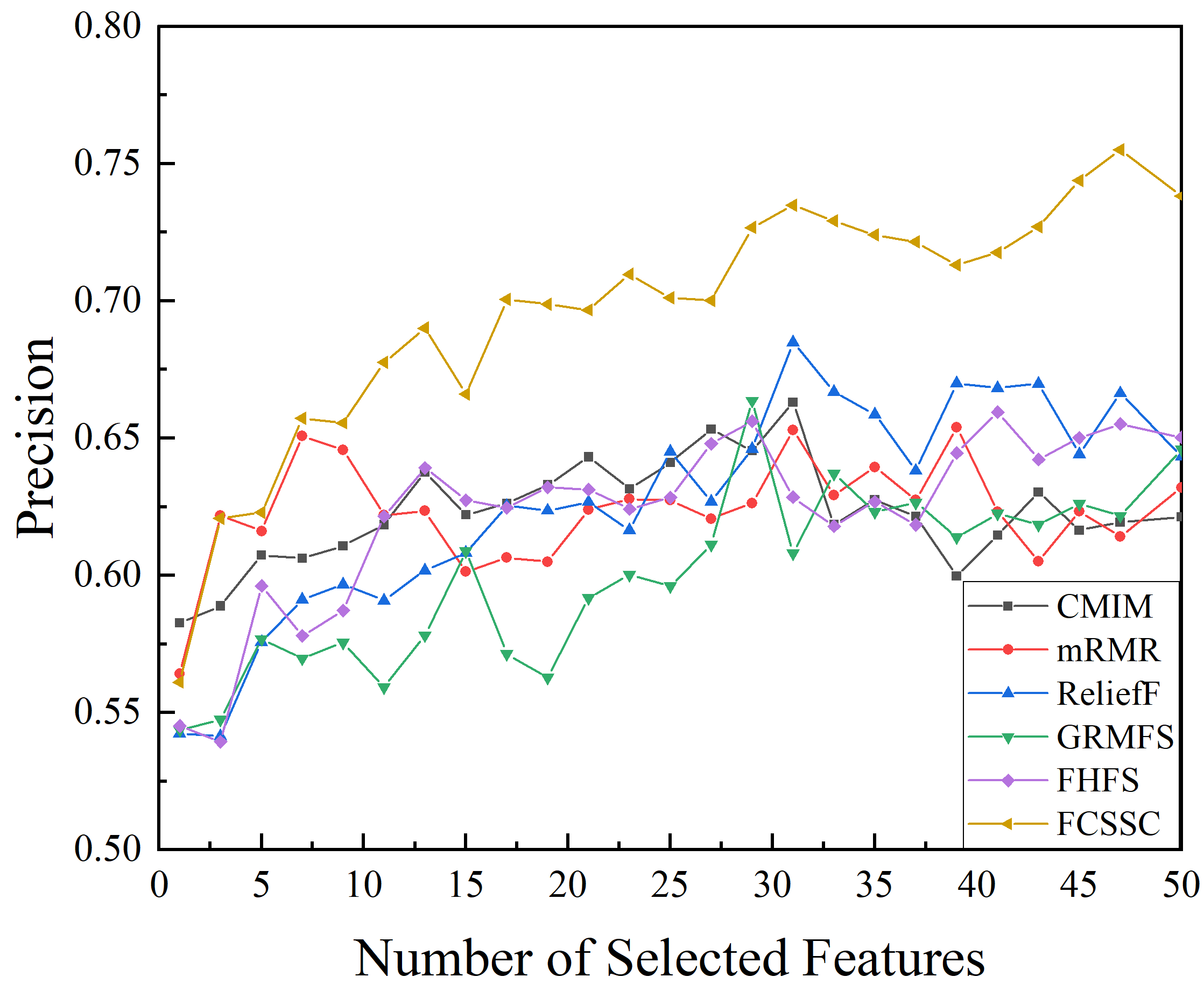}
\label{Fig: precision_features}
\end{minipage}
}
\subfigure[F1-score]{
\centering
\begin{minipage}[t]{0.3\textwidth}
\includegraphics[width=\textwidth]{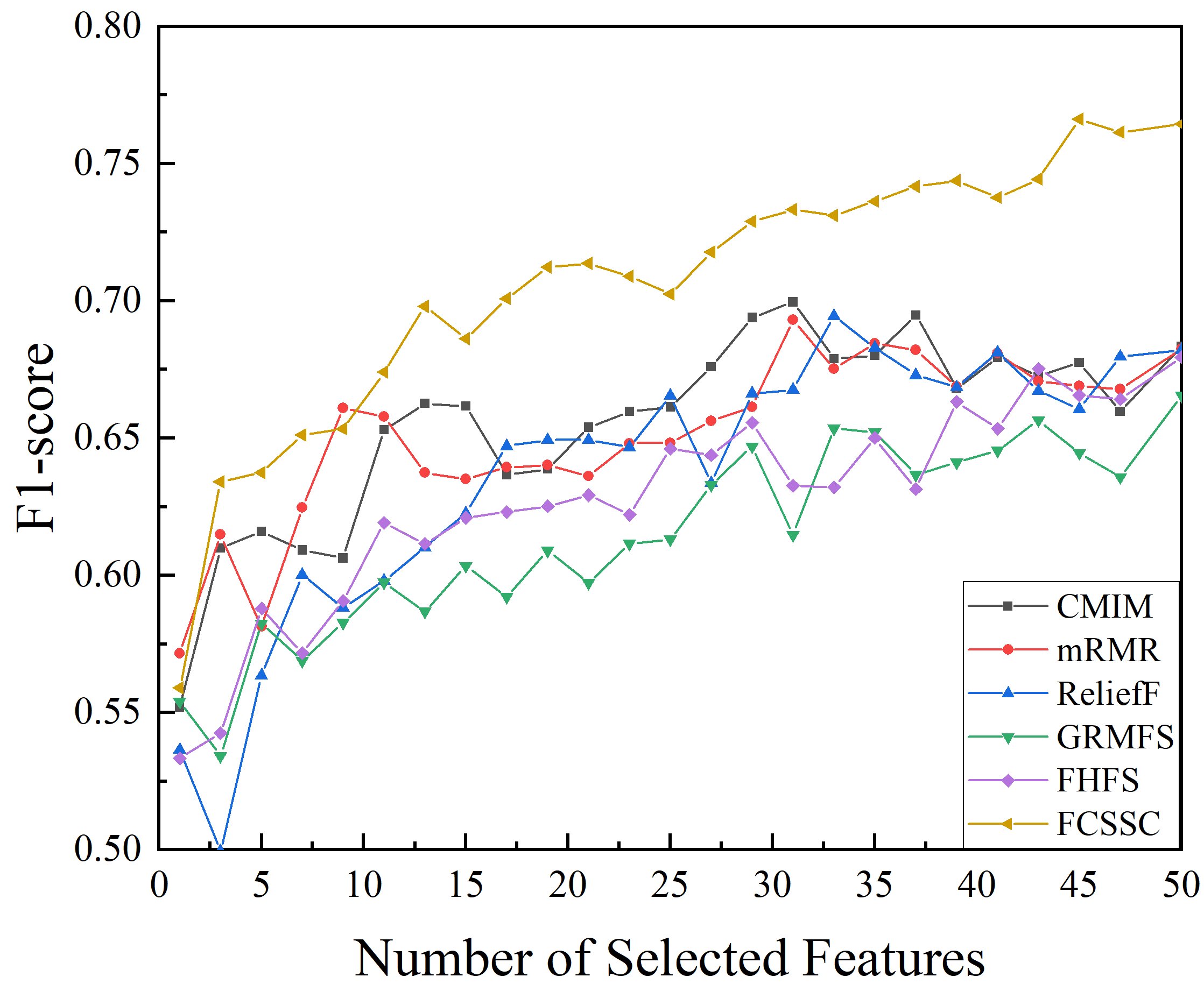}
\label{Fig: f1_features}
\end{minipage}
}
\caption{Classification performance with different features under the KNN classifier in three metrics.}
\label{Fig: Brian_accuracy_features}
\end{figure*}

Figure \ref{Fig: Brian_accuracy_features} represents the performance of classification under the KNN classifier in three metrics including accuracy, precision, and F1-score. The data in Figure \ref{Fig: accuracy_features} show that as the number of selected features increases, there is a noticeable improvement in accuracy across various methods. Notably, the FCSSC method consistently outperforms other methods, showcasing the highest accuracy.
The other methods exhibit fluctuating accuracy but increase as more features are selected. When examining specific cases, when the feature count reaches 50, CMIM, mRMR, and ReliefF exhibit lower accuracy than in situations with fewer than 30 features. This suggests a potential redundancy in feature selection, leading to decreased accuracy.

\begin{figure*}[htbp]
\centering
\subfigure[KNN]{
\centering
\begin{minipage}[t]{0.3\textwidth}
\includegraphics[width=0.9\textwidth]{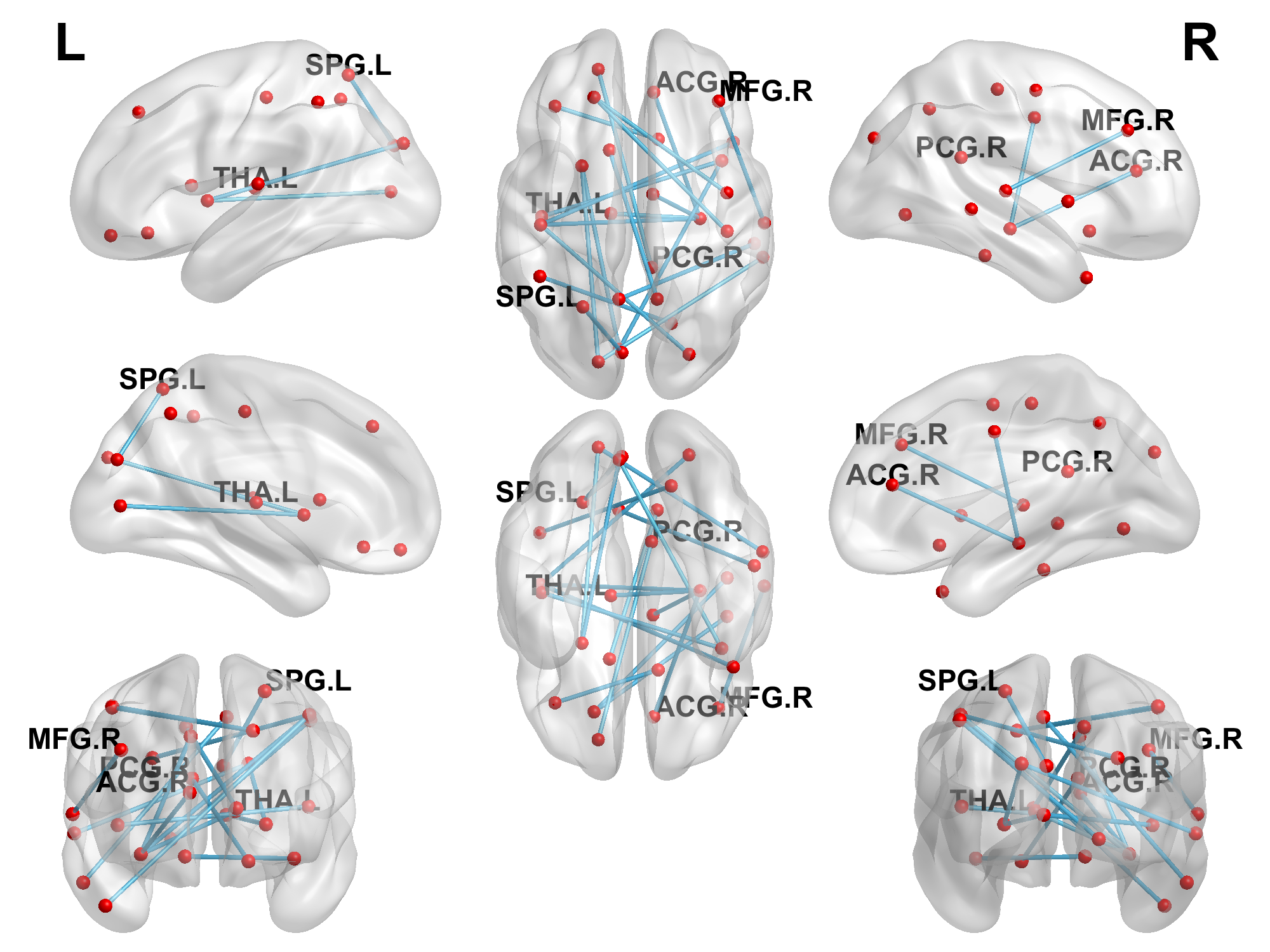}
\label{Fig: edge_knn}
\end{minipage}
}
\subfigure[SVM]{
\centering
\begin{minipage}[t]{0.3\textwidth}
\includegraphics[width=0.9\textwidth]{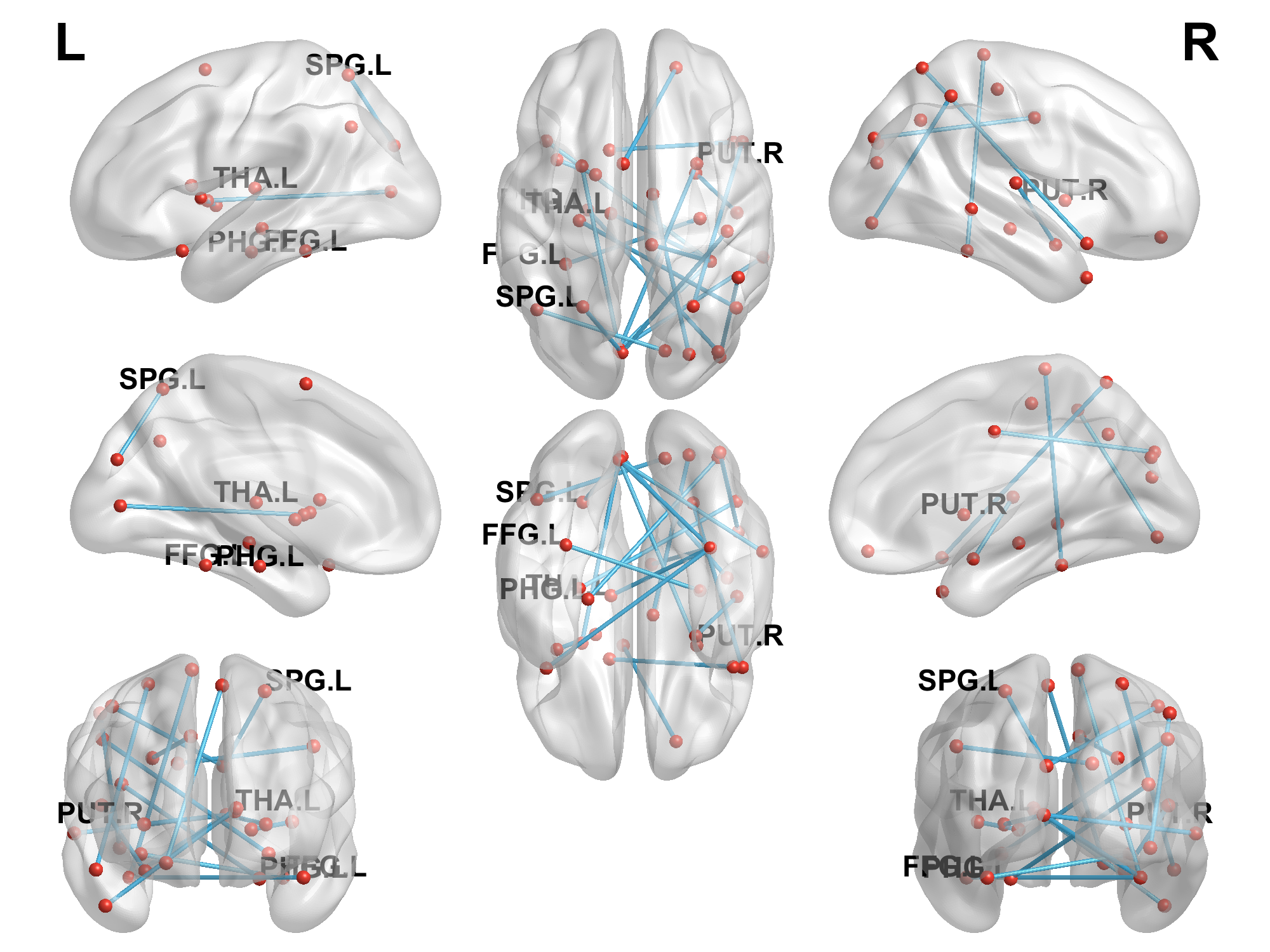}
\label{Fig: edge_svm}
\end{minipage}
}
\subfigure[CART]{
\centering
\begin{minipage}[t]{0.3\textwidth}
\includegraphics[width=0.9\textwidth]{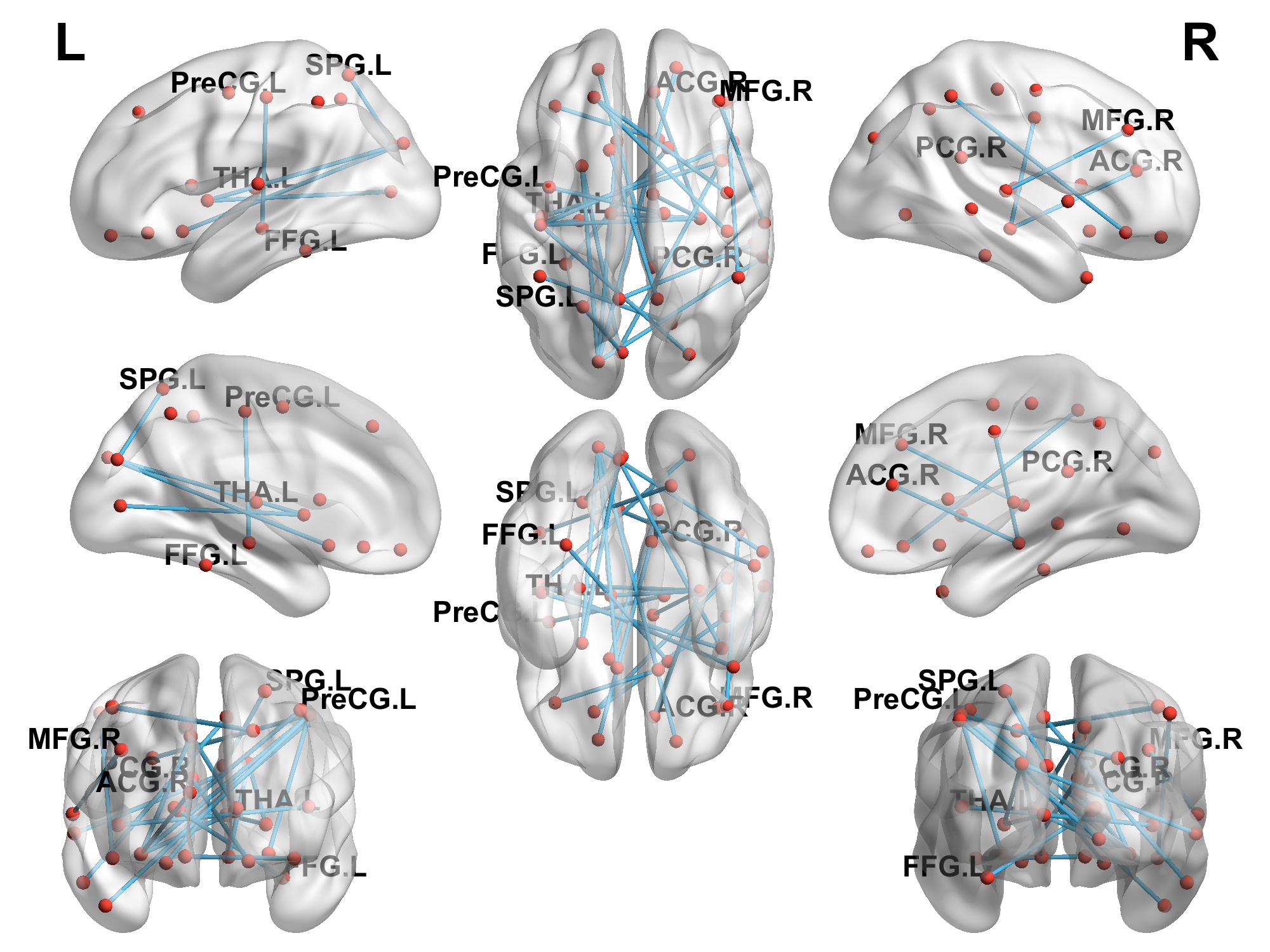}
\label{Fig: edge_cart}
\end{minipage}
}
\caption{Visualization of the top 20 alterations of connectivities differentiating schizophrenia and normal groups under three classifiers.}
\label{Fig: visualization_edge}
\end{figure*}

\begin{figure*}[htbp]
\centering
\subfigure[KNN]{
\centering
\begin{minipage}[t]{0.3\textwidth}
\includegraphics[width=0.9\textwidth]{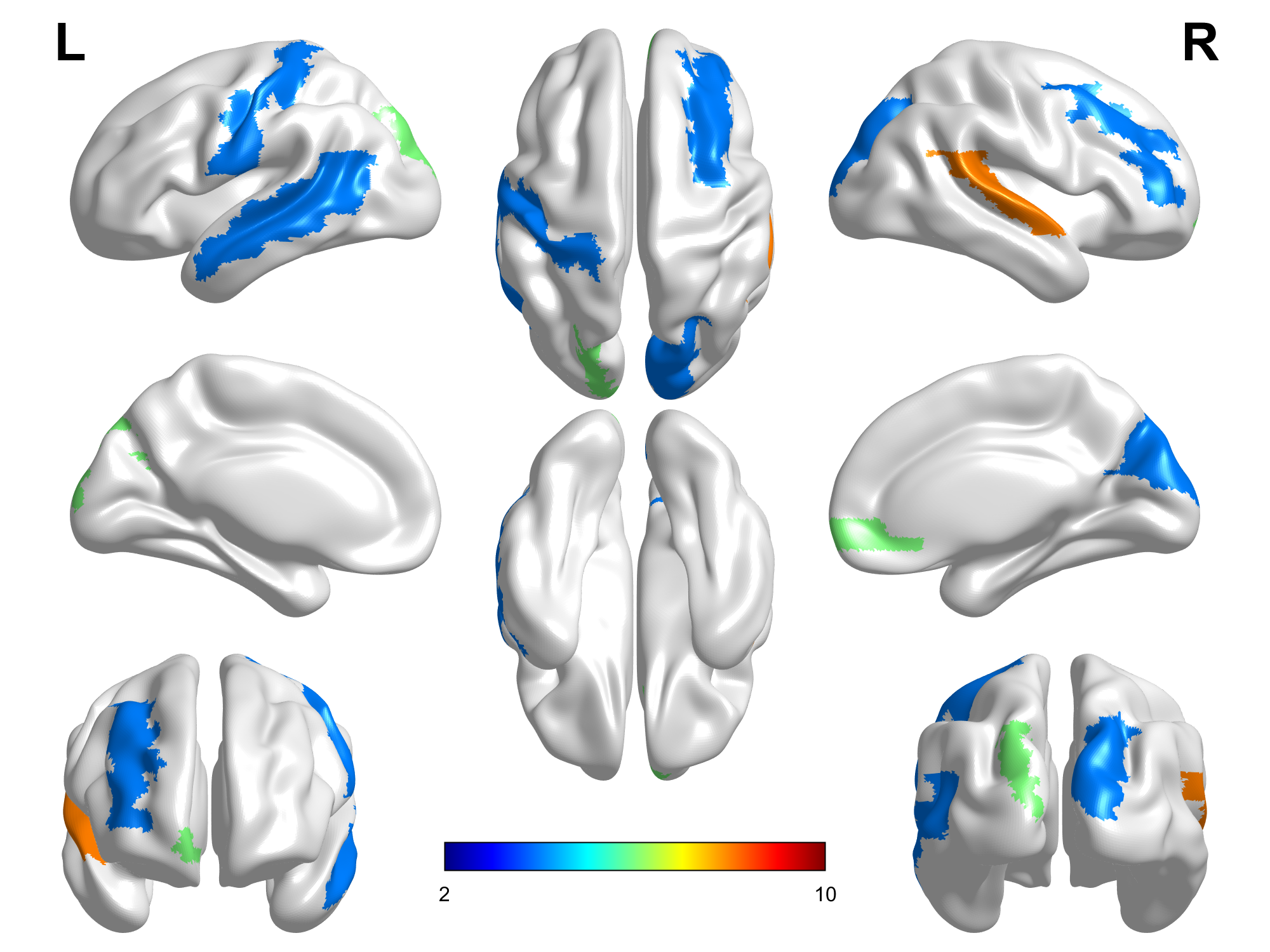}
\label{Fig: region_knn}
\end{minipage}
}
\subfigure[SVM]{
\centering
\begin{minipage}[t]{0.3\textwidth}
\includegraphics[width=0.9\textwidth]{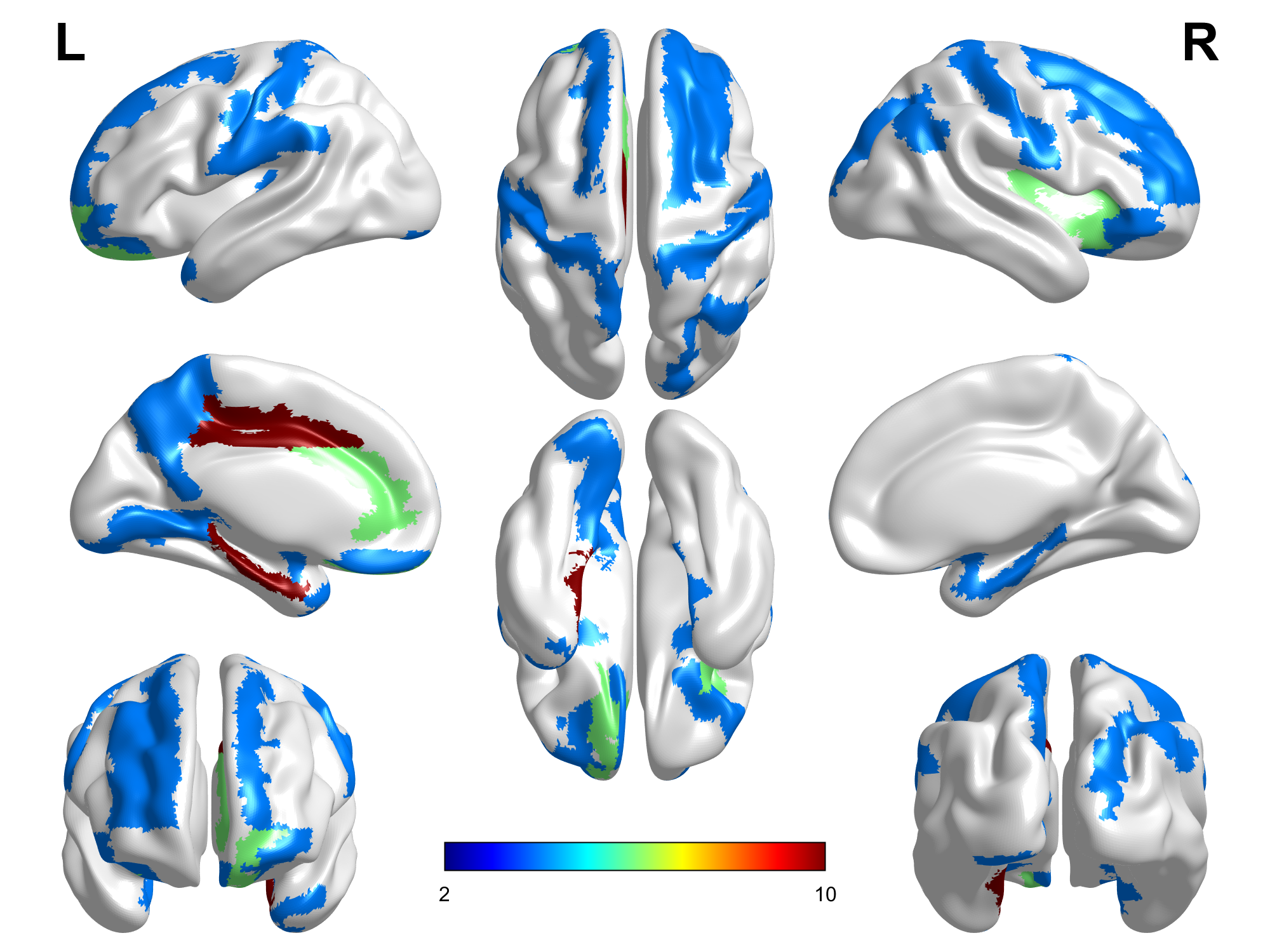}
\label{Fig: region_svm}
\end{minipage}
}
\subfigure[CART]{
\centering
\begin{minipage}[t]{0.3\textwidth}
\includegraphics[width=0.9\textwidth]{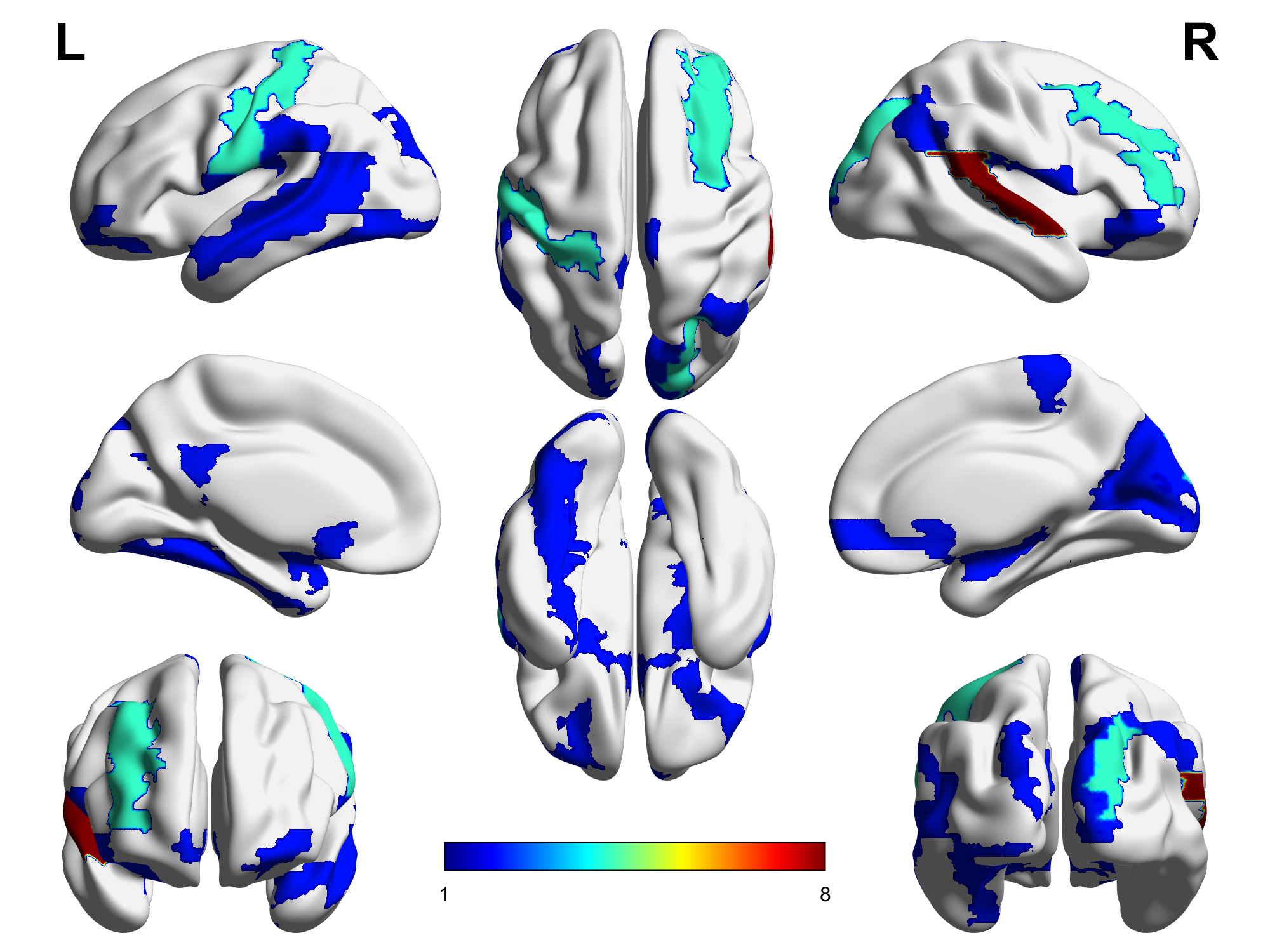}
\label{Fig: region_cart}
\end{minipage}
}
\centering
\caption{Visualization of brain regions differentiating schizophrenia and normal groups under three classifiers.}
\label{Fig: visualization_region}
\end{figure*}

The topological connection analysis of the brain network revealed disparities between the patient and normal groups under three classifiers, as depicted in Figure \ref{Fig: visualization_edge}, showing the most important 20 connectivities. Analyzing these crucial connections reveals specific brain regions that repeatedly appear, indicating their pivotal role in distinguishing between the two groups. To evaluate the significance of brain regions, we sum up the occurrences of brain regions selected for connections under three classifiers, as shown in Figure \ref{Fig: visualization_region}. Extensive literature \cite{V_schizophrenia, V_schizophrenia2, V_schizophrenia3} reinforces the effectiveness of brain regions pinpointed by the FCSSC method in discerning schizophrenia patients from healthy individuals. This result provides strong support for further research and application of brain region identification in schizophrenia. This result strongly supports further research and the application of brain region identification in schizophrenia.

\section{Conclusion} \label{IV}
In this paper, we proposed a cascaded two-stage feature clustering and selection framework that considers both global separability and local consistency of fuzzy decision systems.
The first phase is the feature clustering stage, which groups relevant features into clusters based on their similarities, reducing the search space and the computational complexity. The second phase is a feature selection stage, which selects the most important feature from each cluster based on the fusion of global and local scores, considering the relationships among irrelevant features. The global separability measures the intra-class cohesion and inter-class separation of the features using fuzzy membership concerning the decision class, while the local consistency measures the local consistency of samples with the fuzzy neighborhood rough set model.
We conducted experiments on 18 public datasets and a real-world schizophrenia dataset, evaluating our method against six state-of-the-art feature selection algorithms. The results show that our method achieved superior classification performance while utilizing fewer selected features than the compared algorithms.
Particularly, applying our method to the schizophrenia dataset revealed crucial brain regions and specific characteristics pivotal for diagnosing schizophrenia. This capability showcases the potential for early detection and intervention in schizophrenia cases.

Despite these promising results, there are still several limitations in our work. Firstly, the feature clustering method within our framework plays a crucial role but requires manual setting of the number of clusters. Additionally, the proposed method only considers Euclidean distance between features as a measure of similarity, potentially overlooking complex nonlinear relationships within the data. A potential solution could involve employing kernel-based or deep learning-based methods to learn relationships between features better and perform clustering.
For future work, we plan to extend our method to handle multi-label and multi-view data and explore the use of other clustering and evaluation methods.

\ifCLASSOPTIONcaptionsoff
  \newpage
\fi

\bibliographystyle{IEEEtran}
\bibliography{IEEEabrv,reference.bib}

\begin{IEEEbiography}[{\includegraphics[width=1in,height=1.25in,clip,keepaspectratio]{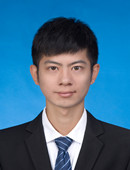}}]{Yuepeng Chen} 
received the B.S. and M.S. degree in communication engineering from the Southeast University, Nanjing, Jiangsu, China, in 2013 and 2016, respectively. He is currently pursuing a doctoral degree at the School of Information Science and Technology, Nantong University, Nantong, China. His main research interests include granular computing, machine learning, and data mining.
\end{IEEEbiography}

\begin{IEEEbiography}[{\includegraphics[width=1in,height=1.25in,clip,keepaspectratio]{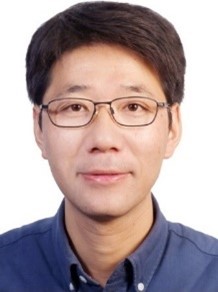}}]{Weiping Ding}
(M’16-SM’19) received the Ph.D. degree in Computer Science, Nanjing University of Aeronautics and Astronautics, Nanjing, China, in 2013. From 2014 to 2015, he was a Postdoctoral Researcher at the Brain Research Center, National Chiao Tung University, Hsinchu, Taiwan, China. In 2016, he was a Visiting Scholar at National University of Singapore, Singapore. From 2017 to 2018, he was a Visiting Professor at University of Technology Sydney, Australia. China. His main research directions involve deep neural networks, granular data mining, and multimodal machine learning. He ranked within the top 2\% Ranking of Scientists in the World by Stanford University (2020-2023). He has published over 300 articles, including over 120 IEEE Transactions papers. His nineteen authored/co-authored papers have been selected as ESI Highly Cited Papers. He has co-authored four books. He has holds 32 approved invention patents, including two U.S. patents and one Australian patent. He serves as an Associate Editor/Area Editor/Editorial Board member of IEEE Transactions on Neural Networks and Learning Systems, IEEE Transactions on Fuzzy Systems, IEEE/CAA Journal of Automatica Sinica, IEEE Transactions on Emerging Topics in Computational Intelligence, IEEE Transactions on Intelligent Transportation Systems, IEEE Transactions on Intelligent Vehicles, IEEE Transactions on Artificial Intelligence, Information Fusion, Information Sciences, Neurocomputing, Applied Soft Computing, Engineering Applications of Artificial Intelligence, Swarm and Evolutionary Computation, et al. He was/is the Leading Guest Editor of Special Issues in several prestigious journals, including IEEE Transactions on Evolutionary Computation, IEEE Transactions on Fuzzy Systems, Information Fusion, Information Sciences, et al. Now he is the Co-Editor-in-Chief of both Journal of Artificial Intelligence and Systems and Journal of Artificial Intelligence Advances.
\end{IEEEbiography}

\begin{IEEEbiography}[{\includegraphics[width=1in,height=1.25in,clip,keepaspectratio]{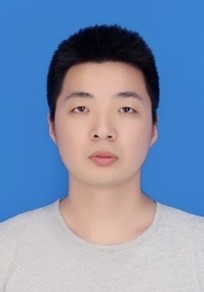}}]{Jiashuang Huang} 
Ph.D., Associate professor. He received his Ph.D. degree in College of Computer Science and Technology from Nanjing University of Aeronautics and Astronautics in 2020. From 2018 to 2019, he was a Visiting Scholar at University of Wollongong (UoW), Wollongong, NSW, Australia. His recent research is to analyze the brain network by using machine learning methods. He has published more than 20 research peer-reviewed journals and conference papers, including IEEE TMI, IEEE JBHI, Medical image Analysis, etc.
\end{IEEEbiography}

\begin{IEEEbiography}[{\includegraphics[width=1in,height=1.25in,clip,keepaspectratio]{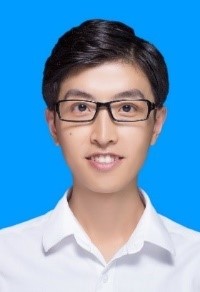}}]
{Hengrong Ju}
PhD, associate professor. He received the B.Sc. and M.Sc. degrees in Computer Science and Technology from Jiangsu University of Science and Technology in 2012 and 2015, respectively, and the Ph.D. degree in Management Science and Engineering from Nanjing University in 2019. From 2017 to 2018, he worked as a visiting scholar at the Department of Electrical and Computer Engineering, University of Alberta. He has authored or co-authored more than 20 scientific papers in international journals and conferences. His current research interests include knowledge discovery and granular computing.
\end{IEEEbiography}

\begin{IEEEbiography}[{\includegraphics[width=1in,height=1.25in,clip,keepaspectratio]{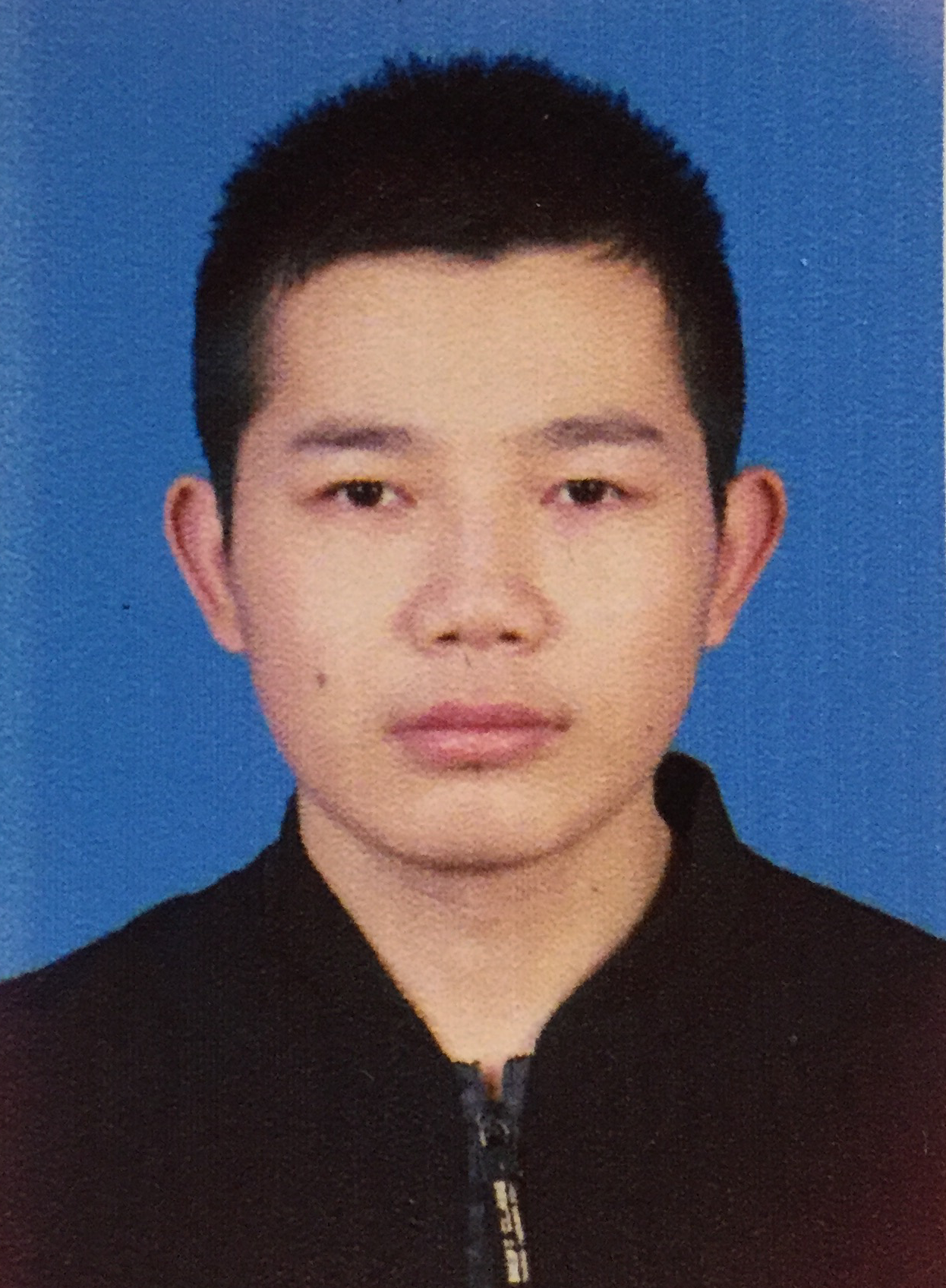}}]
{Tao Yin}
received the B.Sc. degree in Computer Science and Technology from Jiangsu Institute of Technology, Changzhou, China, in 2020. He is currently a doctoral candidate of computer technology in the School of Information Science and Technology, Nantong University. His current research interests include graph convolutional neural network and granular computing.
\end{IEEEbiography}

\end{document}